\pdfoutput=1
\documentclass[11pt]{article}

\usepackage[final]{acl}

\usepackage{tcolorbox}
\usepackage[table]{xcolor}

\usepackage{times}
\usepackage{latexsym}

\usepackage[T1]{fontenc}
\usepackage[utf8]{inputenc}

\usepackage{microtype}

\usepackage{inconsolata}

\usepackage{soul}
\usepackage{url}
\usepackage{graphicx}
\usepackage{amsmath}
\usepackage{amsthm}
\usepackage{booktabs}
\usepackage{algorithm}
\usepackage{algorithmic}
\usepackage[switch]{lineno}
\usepackage{bbding}

\usepackage{multirow}
\usepackage{makecell}
\usepackage{xcolor}
\title{Do Before You Judge:\\Self-Reference as a Pathway to Better LLM Evaluation}
\author{
    Wei-Hsiang Lin\textsuperscript{1} \quad
    Sheng-Lun Wei\textsuperscript{1} \quad
    Hen-Hsen Huang\textsuperscript{2}\quad
    Hsin-Hsi Chen\textsuperscript{1,3}
    \\
    \textsuperscript{1}Department of Computer Science and Information Engineering\\ National Taiwan University, Taiwan
    \\
    \textsuperscript{2}Institute of Information Science, Academia Sinica, Taiwan
    \\
    \textsuperscript{3}AI Research Center (AINTU), National Taiwan University, Taiwan
    \\
    \texttt{\{whlin,weisl\}@nlg.csie.ntu.edu.tw}
    \\
    \texttt{hhhuang@iis.sinica.edu.tw\quad hhchen@ntu.edu.tw}
}

\begin{document}

\maketitle

\begin{abstract}
LLM-as-Judge frameworks are increasingly popular for AI evaluation, yet research findings on the relationship between models' generation and judgment abilities remain inconsistent. We investigate this relationship through systematic dataset- and instance-level analyses across 11 models and 21 diverse tasks. Despite both capabilities relying on the same underlying knowledge, our analyses reveal they are only weakly correlated, primarily due to LLMs' sensitivity to the responses being judged. To address this, we propose a self-reference-guided evaluation strategy that leverages a model's own answers as references. This approach significantly strengthens the correlation between generation and judgment abilities, offering a practical path to align these skills and providing a reliable proxy for model selection in evaluation tasks.
\end{abstract}

\section{Introduction}
Model-based evaluation, which uses large language models (LLMs) as judges, has gained increasing prominence in natural language processing.
This approach, commonly known as LLM-as-Judge \citep{NEURIPS2023_91f18a12}, has been widely adopted across a range of applications \citep{lin-chen-2023-llm, liang-etal-2024-debatrix, fei-etal-2024-lawbench, bi-etal-2024-oceangpt}.
However, a critical question remains: how closely does a model's ability to generate answers correlate with its capacity to evaluate them? Prior work has offered divergent perspectives on this issue, with ~\citet{tan2025judgebench} reporting strong correlation between these abilities and ~\citet{zeng2025mrgsmk} arguing they are not necessarily aligned. These conflicting findings, based only on dataset-level analyses, highlight the need for more comprehensive investigation.

To address this gap, we systematically investigate the relationship between LLMs' answer generation and evaluation capabilities within the LLM-as-Judge framework. We focus on answer judgment as the representative evaluation task, as it directly depends on the same knowledge used in answer generation and thus provides a clearer lens for analyzing their correlation. Given the widespread use of Chain-of-Thought (CoT) reasoning~\citep{NEURIPS2022_9d560961}, we adopt the CoT paradigm for both generation and judgment tasks. Furthermore, unlike prior studies that primarily examine dataset-level correlations, our analysis spans both dataset-level and instance-level perspectives, enabling a more fine-grained understanding of how these capabilities interact.
Further analysis in Section~\ref{sec:5} introduces a self-reference-guided evaluation strategy, which builds upon the reference-guided judging framework~\citep{NEURIPS2023_91f18a12}. This approach significantly improves the correlation between a model's answer generation and judgment capabilities, with an average increase of 0.35 across all evaluated cases (as shown in Table~\ref{tab:self_ref_partial_correlation}).

Our findings reveal that without special techniques, strong answer generation ability does not necessarily translate to strong judgment ability. The correlation between these two capabilities is generally weak when using standard CoT approaches. However, incorporating self-reference-guided judging significantly strengthens this correlation, making generation performance a reliable predictor of judgment capability. These insights offer practical strategies for selecting and utilizing judge models to enhance evaluation performance, effectively aligning generation and judgment capabilities.
Our contributions are fourfold: 
\begin{itemize}
    \item \textbf{Empirical Analysis of LLM Judgment Ability}:
    We demonstrate that LLMs performing well in answer generation do not necessarily excel in answer judgment, highlighting a weak correlation between these abilities under standard evaluation approaches.
    
    \item \textbf{Self-Reference-Guided Judging Effectiveness}:
    Our experiments reveal that incorporating self-reference-guided judging significantly improves the alignment between answer generation and judgment capabilities.
    
    \item \textbf{Practical Implications for Model Selection}:
    Under the self-reference-guided setting, our findings suggest that answer generation ability can serve as a reliable proxy for evaluating judgment capability, enabling more efficient model selection for evaluation tasks.
    
    \item \textbf{Alignment Maintenance Strategy}:
    Our approach supports maintaining alignment between generation and judgment capabilities as LLMs continue to evolve, providing a practical solution to a persistent challenge in LLM evaluation.
\end{itemize}

\section{Related work}
\paragraph{Capabilities of LLM-as-Judge.} Prior work on LLM-as-Judge has explored their capabilities through benchmarks \cite{NEURIPS2023_91f18a12, li2024crowdsourceddatahighqualitybenchmarks, wei2024systematicevaluationllmasajudgellm}, tuning methods \cite{zhu_judgelm_2023, wang2024pandalm, lee-etal-2024-aligning}, prompting strategies \cite{wang-etal-2024-large-language-models-fair, raina-etal-2024-llm, badshah2024referenceguidedverdictllmsasjudgesautomatic}, and model interaction architectures \cite{chan2024chateval, chen-etal-2024-reconcile, verga2024replacingjudgesjuriesevaluating}. Several survey papers \cite{li2025generationjudgmentopportunitieschallenges, li2024llmsasjudgescomprehensivesurveyllmbased} have summarized these developments, evaluating various prompt designs \cite{liu_reife_2024} and implementation approaches.

\paragraph{Limitations of LLM-as-Judge.} Recent research has identified limitations in this framework, particularly biases in model judgments~\citep{wei_unveiling_2024, NEURIPS2023_91f18a12, koo-etal-2024-benchmarking}, with comprehensive overviews provided in survey works~\citep{chen-etal-2024-humans, shi_judging_2024}. A critical yet underexplored aspect is how LLM judges are selected. Models are often chosen based on their generation performance~\citep{hendrycks2021measuring, hendrycksmath2021} or leaderboard rankings~\citep{pmlr-v235-chiang24b}, assuming strong generation capabilities imply strong judgment capabilities. However, studies present conflicting views: Tan et al.~\citep{tan2025judgebench} report strong correlation between these abilities, while Zeng et al.~\citep{zeng2025mrgsmk} argue they are not necessarily aligned. Both rely solely on dataset-level analyses with limited benchmarks. We address this gap by systematically examining the relationship between generation and judgment abilities at both dataset and instance levels, offering insights into when these capabilities diverge and how they can be effectively aligned.

\section{Framework for Evaluating LLM Judgment Ability}

\subsection{Objective \& Notations}\label{Objective}
The goal of this paper is to investigate whether LLMs' ability to evaluate answers correlates with their ability to generate correct answers for the same questions. 
In other words, we aim to determine whether proficiency in ``answering questions" implies proficiency in ``judging answers" (and vice versa) and to analyze the potential correlation between these two competencies. 
The experiment involves two roles: the agent model $\mathcal{M}_A$, which generates answers during the answer generation stage, and the judge model $\mathcal{M}_J$, which evaluates their correctness during the answer judgment stage.

\paragraph{Answer Generation.}
Let $\mathcal{D}_{G} = \{(q_i, a^*_i)\}_{i=1}^N$ be a dataset consisting of $N$ questions, where $q_i$ is the $i$-th question and $a^*_i$ is the ground-truth answer for $q_i$. Let $\mathcal{M}_A$ denote the agent model, for each question $q_i$, the agent model generates an answer:

\begin{equation}
\label{answer-generation}
    \hat{a}_i^{\mathcal{M}_A} = \mathcal{M}_A(q_i)
\end{equation}

\noindent Similarly, we perform the answer generation process for the judge model $\mathcal{M}_J$ to assess its capacity to answer the question:

\begin{equation}
\label{judge-answer-generation}
    \hat{a}_i^{\mathcal{M}_J} = \mathcal{M}_J(q_i)
\end{equation}

\noindent The capacity of the judge model is defined as:

\begin{equation}
\label{acc}
    \mathrm{Acc}_{Generation}^{\mathcal{M}_J}
= \frac{\bigl|\{\,i \;\mid\; \hat{a}_i^{\mathcal{M}_J} = a_i^* \}\bigr|}{N}
\end{equation}

\noindent where $\bigl|\{\,i \;\mid\; \hat{a}_i^{\mathcal{M}_J} = a_i^* \}\bigr|$ represents the number of questions that the judge model $\mathcal{M}_J$ provides the correct answer.

\paragraph{Answer Judgment.}
After the answer generation stage, we proceed to the answer judgment stage. In this stage, we first construct the dataset $\mathcal{D}_J$ and then use the judge model $\mathcal{M}_J$ to evaluate the correctness of the results generated in the previous stage. $\mathcal{D}_J$ is defined as:

\begin{equation}
\label{equation:judge}
\mathcal{D}_J \;=\; \bigl\{\,\bigl(q_i, \hat{a}_i^{\mathcal{M}_A}, y_i^*\bigr)\bigr\}_{i=1}^N
\end{equation}

\noindent where $q_i$ is the $i$-th question from $\mathcal{D}_G$, $\hat{a}_i^{\mathcal{M}_A}$ is the answer generated by model $\mathcal{M}_A$ in Equation (\ref{answer-generation}), and $y_i^*$ indicates whether $\hat{a}_i^{\mathcal{M}_A}$ is correct, defined as:

\begin{equation}
y_i^* =
\begin{cases} 
1, & \text{if } \hat{a}_i^{\mathcal{M}_A} = a_i^* \\
0, & \text{otherwise.}
\end{cases}
\end{equation}

\noindent We then use the judge model to generate judgment results as:

\begin{equation}
y_i^{\mathcal{M}_J} = \mathcal{M}_J(q_i, \hat{a}_i^{\mathcal{M}_A})
\end{equation}

\noindent Finally, we can evaluate the model's judgment capability via:

% \begin{equation}
% \mathrm{P}_{Judge}^{\mathcal{M}_J} = \frac{\bigl|\{\,i \;\mid\; y_i^{\mathcal{M}J} = 1 \;\wedge\; y_i^* = 1 \}\bigr|}{\bigl|\{\,i \;\mid\; y_i^{\mathcal{M}J} = 1 \; \}\bigr|}
% \end{equation}

% \begin{equation}
% \mathrm{R}_{Judge}^{\mathcal{M}_J} = \frac{\bigl|\{\,i \;\mid\; y_i^{\mathcal{M}J} = 1 \;\wedge\; y_i^* = 1 \}\bigr|}{\bigl|\{\,i \;\mid\;y_i^* = 1 \}\bigr|}
% \end{equation}

\begin{equation}
\mathrm{P}_{Judge} =
\frac{\bigl|\{\,i \mid y_i^{\mathcal{M}_J}=1 \land y_i^*=1 \}\bigr|}
     {\bigl|\{\,i \mid y_i^{\mathcal{M}_J}=1 \}\bigr|}
\end{equation}

\begin{equation}
\mathrm{R}_{Judge} =
\frac{\bigl|\{\,i \mid y_i^{\mathcal{M}_J}=1 \land y_i^*=1 \}\bigr|}
     {\bigl|\{\,i \mid y_i^*=1 \}\bigr|}
\end{equation}

\begin{equation}
\label{f1}
\mathrm{F1}_{Judge} = 2 \times \frac{\mathrm{P}_{Judge}\times \mathrm{R}_{Judge}}{\mathrm{P}_{Judge} + \mathrm{R}_{Judge}}
\end{equation}

\paragraph{Correlation.}\label{Correlation}
To quantify the linear relationship between the \emph{answer generation ability} and the \emph{answer judgement ability} of the judge model \(\mathcal{M}_J\) at the instance level, we introduce three binary variables for each question \(i\):

\noindent The event that the judge model \(\mathcal{M}_J\) answers \(q_i\) correctly is defined as
\begin{equation}
\label{gi}
G_i = \mathbf{1}\!\bigl[\hat{a}_i^{\mathcal{M}_J}=a_i^*\bigr]
\end{equation}
\noindent The event that the judge model correctly classifies the agent's answer is defined as
\begin{equation}
\label{ji}
J_i = \mathbf{1}\!\bigl[y_i^{\mathcal{M}_J} = y_i^*\bigr]
\end{equation}
\noindent The event that the agent model \(\mathcal{M}_A\) answers \(q_i\) correctly is defined as
\begin{equation}
\label{ai}
A_i = y_i^* = \mathbf{1}\!\bigl[\hat{a}_i^{\mathcal{M}_A}=a_i^*\bigr]
\end{equation}
Using the \(N\) triplets \(\{(G_i,J_i,A_i)\}_{i=1}^N\), we first compute the pairwise Pearson correlation coefficients:
\begin{equation}
\begin{aligned}
r_{G,J} &= \operatorname{corr}(G,J),\quad
r_{G,A} = \operatorname{corr}(G,A)
\\
r_{J,A} &= \operatorname{corr}(J,A)
\end{aligned}
\end{equation}
\vspace{4pt}
\noindent
% \textbf{Partial correlation.}  
% As our experiments will show (Section \ref{sec:4_2}), we discovered that the correctness of the agent's response significantly influences judgment performance. To control for this confounding effect and observe the true correlation between answer generation and judgment abilities, we employ partial correlation analysis. The partial correlation between \(G\) and \(J\) given \(A\) removes the linear influence of \(A\) (i.e., the correctness of the evaluated response) from both variables:

\noindent \textbf{Partial Correlation.}  
We hypothesize that the correctness of the agent's response may significantly influence judgment performance, potentially acting as a confounding factor in our analysis. We will directly investigate this potential influence in our subsequent analysis. To control for this possible effect and observe the underlying correlation between answer generation and judgment abilities, we employ partial correlation analysis. The partial correlation between \(G\) and \(J\) given \(A\) removes the linear influence of \(A\) (i.e., the correctness of the evaluated response) from both variables:
\begin{equation}
\label{eq:partial_corr}
r_{G,J\mid A} =
\frac{r_{G,J}-r_{G,A}\,r_{J,A}}
     {\sqrt{\bigl(1-r_{G,A}^2\bigr)\bigl(1-r_{J,A}^2\bigr)}}
\end{equation}
\noindent
Here, \(r_{G,J\mid A}=0\) indicates no residual linear association between the judge model's generation accuracy and judgement accuracy once the agent model's correctness is held fixed, while \(|r_{G,J\mid A}|\) approaching~1 signals a strong intrinsic link between these two competencies that is \emph{not} attributable to the quality of the agent's answer. This approach allows us to assess whether the correlation between generation and judgment capabilities exists independently of the evaluated response quality, addressing limitations in prior studies that relied solely on dataset-level correlations. 

% In our subsequent analysis, we will directly investigate if the correctness of the agent's response influences judgment performance, validating our choice of analytical approach.

% This approach allows us to assess whether the correlation between generation and judgment capabilities exists independently of the evaluated response quality, addressing limitations in prior studies that relied solely on dataset-level correlations.

%% --------------------------- Analysis 1 figure --------------------------------------
\begin{figure*}[ht]
    \centering
    \includegraphics[width=0.95\textwidth]{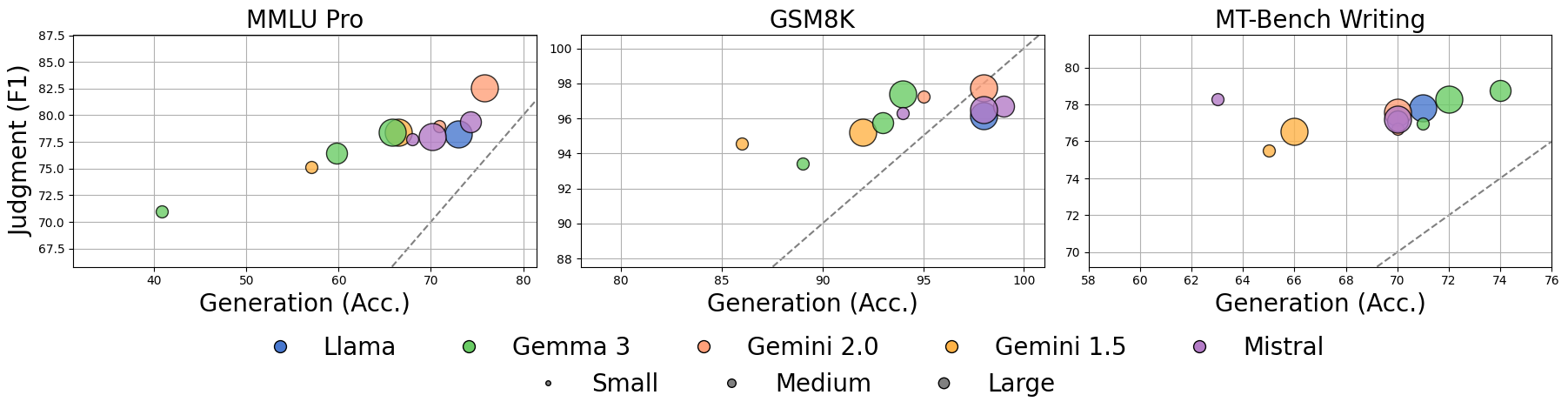} % 圖片文件的路徑
    \caption{Relationship between the capabilities of answer generation (measured by accuracy) and answer judgment (measured by F1 score) across three datasets: MMLU Pro, GSM8K, and MT-bench writing. Each subplot corresponds to one dataset. Different colors represent different model series, and the size of each circle reflects the relative size of models within the same series.}
    \label{fig:capability_comparison}
\end{figure*}

%% --------------------------- Analysis 1 figure --------------------------------------

\subsection{Evaluation Models}
We adopt relatively weaker LLMs as agent models and relatively stronger LLMs as judge models. 
For the agent models, we include Ministral 8B~\citep{ai_ministral_2024} and Llama 3.1 series~\citep{dubey2024llama}. 
For the judge models, we consider four different families of LLMs, including Llama 3.1 405B, the Mistral series~\citep{ai_large_2024}, the Gemini series~\citep{Anil2023GeminiAF,gemini15}, and the Gemma series~\citep{gemmateam2025gemma3technicalreport}. Detailed model information and implementation details are provided in Appendix~\ref{appendix:model_info}.

% For additional implementation details, for Llama 3.1, we leverage the API provided by SambaNova\footnote{\url{https://cloud.sambanova.ai}} to optimize the efficiency and scalability of our experiments. 
% For the Mistral series, Gemini series, and Gemma series, we utilize APIs provided by Mistral and Google. Due to space constraints, detailed model information is provided in Appendix~\ref{appendix:model_info}.

\subsection{Evaluation Tasks}

Our experiments cover seven datasets that span multiple answer formats and domains, comprising a total of 21 subtasks.
For multiple-choice questions (MCQs), we use MMLU Pro~\cite{NEURIPS2024_ad236edc}.
For dialogue tasks with human-preference annotations, we rely on Chatbot Arena~\cite{NEURIPS2023_91f18a12} and three subsets of MT-Bench~\cite{NEURIPS2023_91f18a12}: \emph{Humanities}, \emph{Roleplay}, and \emph{Writing}, which add open-ended Q\&A, scenario-based role-playing, and creative-writing challenges within the same dialogue-and-preference framework.
To examine mathematical and symbolic reasoning, we include GSM8K~\cite{cobbe_training_2021} and GSM-Symbolic (P1 and P2)~\cite{mirzadeh2025gsmsymbolic}.

For each task, we randomly sample 100 instances.
In the case of MMLU Pro, which contains 14 subtasks, we sample 100 instances per subtask, yielding 1,400 MMLU Pro examples in total.
Since we use three different agent models (Ministral 8B, Llama 3.1 8B, and Llama 3.1 70B) to generate answers during the judgment phase, our judgment evaluation dataset effectively triples in size, providing robust coverage across different answer quality distributions.
Because of space limits, the main text reports results on three representative datasets, namely MMLU Pro, GSM8K, and MT-Bench Writing, which together capture the trends observed across the full benchmark suite. Complete results for the remaining four datasets appear in Appendix \ref{appendix:full_results}.
Together, these seven datasets allow us to probe LLM judges across MCQs, dialogue with human preferences, creative writing, and mathematical reasoning tasks.

\subsection{Limitations of Pairwise Evaluation}
% Previous LLM-as-Judge research typically employs pairwise evaluation, where judge models select the better result from multiple outputs generated by agent models. 
% However, they focus on comparing the correctness of multiple options rather than directly investigating the relationship between answer generation and the ability to judge.
% To design experiments that more clearly reflect these two abilities, we adopt a pointwise setup as described in Section~\ref{Objective}. 
Prior work on LLM-as-Judge primarily adopts a \textit{pairwise} evaluation paradigm, where a judge model selects the better output among multiple candidates generated by agent models. However, such approaches emphasize comparative correctness across options, rather than directly disentangling the ability to \emph{generate} answers from the ability to \emph{judge} them. To more clearly isolate and evaluate these two capabilities, we adopt a \textit{pointwise} setup (Section~\ref{Objective}). Moreover, the pointwise setting mitigates potential selection bias~\citep{wei_unveiling_2024, NEURIPS2023_91f18a12, koo-etal-2024-benchmarking} inherent in pairwise evaluation, where the LLM is forced to solve a multiple-choice style problem that may confound its true judging ability.

\begin{table*}[ht]
\centering
\small
\setlength{\tabcolsep}{4pt}
\renewcommand{\arraystretch}{1}
\begin{tabular}{lccccccccc}
\hline
\multirow{2}{*}{\textbf{Judge model}}
 & \multicolumn{3}{c}{\textbf{GSM8K}}
 & \multicolumn{3}{c}{\textbf{MMLUPro}}
 & \multicolumn{3}{c}{\textbf{MT-Bench Writing}}
 \\
\cline{2-10}
 & \Checkmark & \XSolidBrush & $\Delta$ & \Checkmark & \XSolidBrush & $\Delta$ & \Checkmark & \XSolidBrush & $\Delta$ \\
\hline
Llama 3.1 405B & \textbf{96.85} & 0.00 & \textcolor{blue}{96.85} & \textbf{88.75} & 32.74 & \textcolor{blue}{56.01} & \textbf{93.51} & 9.41 & \textcolor{blue}{84.10} \\
Gemini 2.0 Flash & \textbf{98.13} & 0.00 & \textcolor{blue}{98.13} & \textbf{90.84} & 35.31 & \textcolor{blue}{55.53} & \textbf{91.11} & 28.28 & \textcolor{blue}{62.83} \\
Gemini 2.0 Flash Lite & \textbf{98.68} & 33.33 & \textcolor{blue}{65.35} & \textbf{89.33} & 36.77 & \textcolor{blue}{52.56} & \textbf{90.77} & 23.76 & \textcolor{blue}{67.01} \\
Gemini 1.5 Flash & \textbf{97.10} & 52.17 & \textcolor{blue}{44.93} & \textbf{89.56} & 38.07 & \textcolor{blue}{51.49} & \textbf{90.30} & 34.86 & \textcolor{blue}{55.44} \\
Gemini 1.5 Flash 8B & \textbf{98.16} & 65.57 & \textcolor{blue}{32.59} & \textbf{88.43} & 43.81 & \textcolor{blue}{44.62} & \textbf{87.76} & 40.68 & \textcolor{blue}{47.08} \\
Gemma 3 4B & \textbf{96.58} & 47.06 & \textcolor{blue}{49.52} & \textbf{86.41} & 54.89 & \textcolor{blue}{31.52} & \textbf{85.39} & 50.45 & \textcolor{blue}{34.94} \\
Gemma 3 12B & \textbf{97.90} & 33.33 & \textcolor{blue}{64.57} & \textbf{88.41} & 46.75 & \textcolor{blue}{41.66} & \textbf{87.89} & 45.36 & \textcolor{blue}{42.53} \\
Gemma 3 27B & \textbf{98.28} & 70.59 & \textcolor{blue}{27.69} & \textbf{89.41} & 43.59 & \textcolor{blue}{45.82} & \textbf{89.19} & 38.00 & \textcolor{blue}{51.19} \\
Mistral Small 3.1 & \textbf{97.74} & 42.86 & \textcolor{blue}{54.88} & \textbf{88.51} & 39.23 & \textcolor{blue}{49.28} & \textbf{94.74} & 30.51 & \textcolor{blue}{64.23} \\
Mistral Medium 3 & \textbf{97.04} & 50.00 & \textcolor{blue}{47.04} & \textbf{88.87} & 38.19 & \textcolor{blue}{50.68} & \textbf{91.92} & 24.00 & \textcolor{blue}{67.92} \\
Mistral Large 2 & \textbf{96.63} & 90.91 & \textcolor{blue}{5.72} & \textbf{89.04} & 35.34 & \textcolor{blue}{53.70} & \textbf{92.27} & 17.02 & \textcolor{blue}{75.25} \\
\hline
\end{tabular}
\caption{Answer judgment performance (\%) across different models and datasets. \Checkmark represents $\mathcal{D}_{J+}$, and \XSolidBrush represents $\mathcal{D}_{J-}$. \(\Delta\) denotes the gap between the performance of $\mathcal{D}_{J+}$ and $\mathcal{D}_{J-}$ for the same model and dataset. If the performance of $\mathcal{D}_{J+}$ is higher, it is marked in {\color{blue}{blue}}; otherwise, it is marked in {\color{red}{red}}.}
\label{tab:full_analysis2}
\end{table*}

\section{Relationship between Answer Generation and Answer Judgment}
\label{sec:4}
\subsection{Dataset-Level Observations}
Following the process described in Section~\ref{Objective}, we evaluate the capabilities of answer generation and answer judgment using Equation~\ref{acc} and Equation~\ref{f1}, respectively.  Figure~\ref{fig:capability_comparison} plots these two metrics for eleven models on three representative datasets, chosen for space constraint and because they typify the trends observed on the full benchmark suite.  
Complete results for the remaining four datasets are reported in Appendix~\ref{appendix:full_results}.

% The results across three datasets and eleven models are shown in Figure~\ref{fig:capability_comparison} to illustrate the relationship between these two capabilities. Additional results for the remaining datasets are provided in Appendix~\ref{....}.

% The results indicate that MMLU Pro is the most challenging dataset among the four, with an accuracy gap of over 15\% across all models. Additionally, we divide the MMLU Pro dataset into 14 subtasks to investigate whether this relationship varies across different domains. 

The results, shown in Figure~\ref{fig:capability_comparison}, reveal a clear positive correlation: models with higher answer generation performance typically exhibit better answer judgment performance across all datasets. Although the relationship is not strictly linear, a discernible trend indicates that superior answer generation capabilities generally correspond to enhanced answer judgment performance. This finding aligns with the observations in \citet{tan2025judgebench}; however, we do not observe the performance drop in answer evaluation compared to generation reported by \citet{zeng2025mrgsmk}. This discrepancy may be attributed to our evaluation task focusing on binary correctness judgment, which naturally yields higher expected performance.

We further investigate the primary reason behind this observation. Specifically, we seek to distinguish whether this correlation is rooted in the shared knowledge required for both generation and judgment on a given task, or if it is an artifact of stronger models' generally superior performance across various tasks.
% Specifically, we aim to determine whether this correlation arises because the knowledge required for generating answers aligns with that required for judging answers, or if it is simply due to stronger models performing better across different tasks, suggesting that the correlation does not necessarily imply causation. 
To clarify this relationship, we conduct further analysis and address the following question: "When the judge model correctly answers a question, does it judge other models' responses more accurately?"
% When the judge model correctly answers a question, does it judge other models' responses to the same question more accurately?

To investigate whether the internal knowledge of the judge model affects its performance in evaluation tasks, we split the dataset $\mathcal{D}_J$, as defined in Equation~\ref{equation:judge}, into two subsets: $\mathcal{D}_{J+}$ and $\mathcal{D}_{J-}$. The former represents samples where the judge model $\mathcal{M_J}$ answered correctly, while the latter contains samples where it failed to provide the correct answer. Note that for different judge models $\mathcal{M_J}$, the split datasets $\mathcal{D}_{J+}$ and $\mathcal{D}_{J-}$ are distinct. This experimental design isolates the answer generation capability of the judge model, enabling more precise examinations of how the model's internal knowledge impacts its judging effectiveness.

Table~\ref{tab:full_analysis2} shows the performance of answer judgment on $\mathcal{D}_{J+}$ and $\mathcal{D}_{J-}$ across eleven different models. In all cases , the judge models achieve significantly higher F1 scores on $\mathcal{D}_{J+}$. This observation appears to suggest that the evaluation ability of judge models is more effective when they possess related knowledge, as indicated by their ability to answer the question correctly. 

% Furthermore, we conduct a deeper analysis on 14 MMLU Pro subtasks. As shown in Figure~\ref{fig}, across all subtasks and models, the performance of answer judgment in $\mathcal{D}{J+}$ consistently outperforms that in $\mathcal{D}_{J-}$, aligning with the findings presented in Table~\ref{tab}.

% While these dataset-level observations provide valuable insights, they may not fully explain the observed correlation between answer generation and answer judgment abilities. The superior performance of judges on $\mathcal{D}_{J+}$ raises further questions:
% \begin{itemize}
% \item Does this truly indicate a causal relationship between answer generation ability and answer judgment ability?
% \item Could $\mathcal{D}_{J+}$ and $\mathcal{D}_{J-}$ possess other inherent characteristics that contribute to these performance differences?
% \item How can we reconcile the conflicting conclusions from previous works regarding the correlation between answer generation and answer judgment capabilities?
% \end{itemize}

% While these dataset-level observations offer useful insights, they may not fully account for the observed correlation between answer generation and judgment capabilities. The strong performance of judges on $\mathcal{D}_{J+}$ raises several questions:

While these dataset-level observations offer useful insights, they provide insufficient evidence to draw definitive conclusions about the correlation between answer generation and judgment capabilities. The strong performance of judges on $\mathcal{D}_{J+}$ raises several questions:
\begin{itemize}
% \item Does this imply a strong correlation between generation and judgment abilities?
\item Does this imply that models with strong generation abilities will have strong judgment abilities, suggesting these capabilities are highly correlated?
\item Could inherent differences between $\mathcal{D}_{J+}$ and $\mathcal{D}_{J-}$ explain the large judgment performance gap, rather than ability correlation?
% \item Could $\mathcal{D}_{J+}$ and $\mathcal{D}_{J-}$ differ in other ways that affect performance?
\item How can we reconcile prior conflicting findings \cite{tan2025judgebench,zeng2025mrgsmk} regarding the relationship between these two capabilities?
\end{itemize}

\begin{figure*}[tp]
    \centering
    \includegraphics[width=1\textwidth, height=7.5cm]{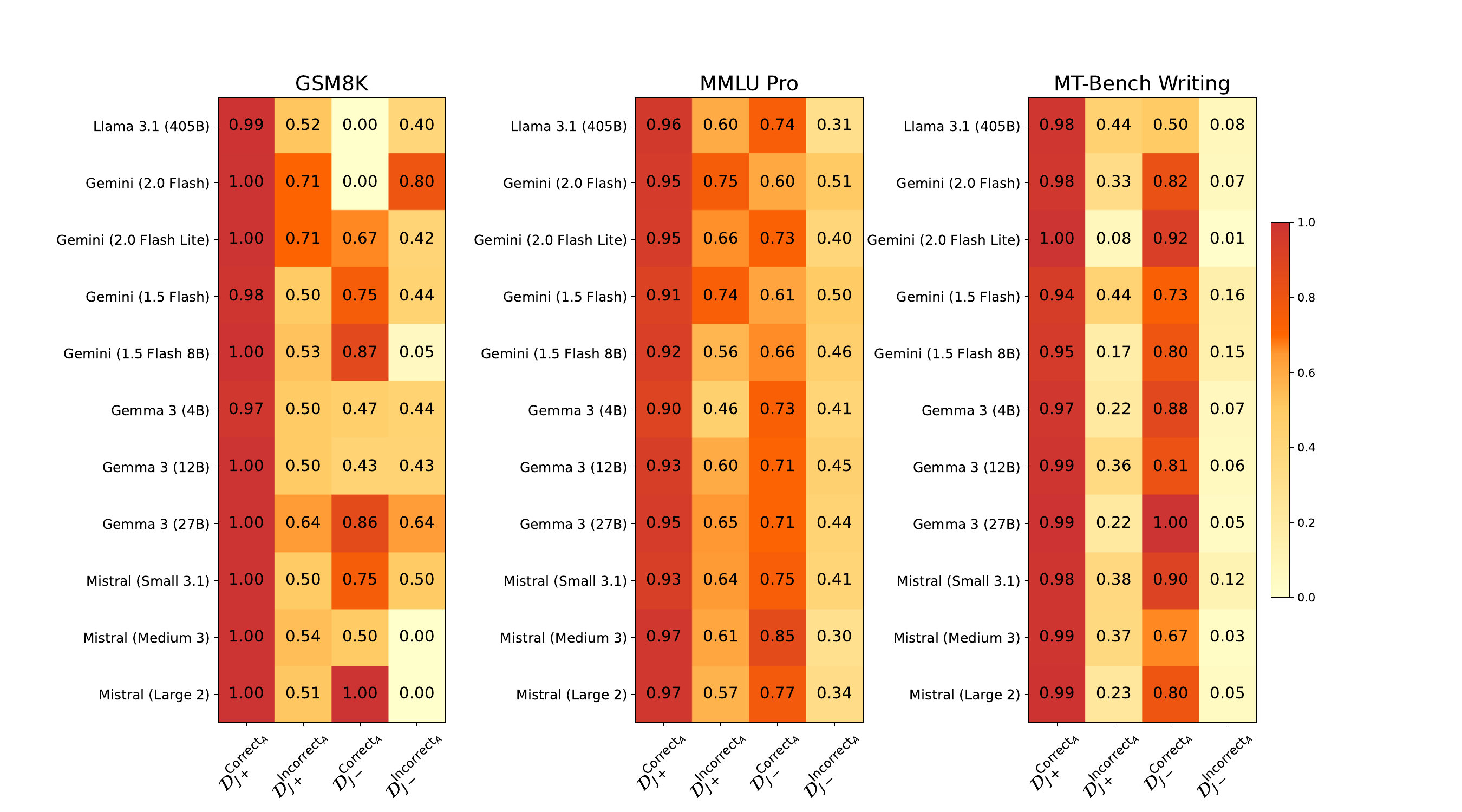} % 圖片文件的路徑
    \caption{Heatmap Visualization of Evaluation Performance across Datasets. This figure illustrates the integration of the judge model's answer generation capabilities with the labels of evaluation questions across four datasets. Each dataset is represented in one of four subfigures, with subsets $\mathcal{D}_{J+}^{\text{Correct}_A}$, $\mathcal{D}_{J+}^{\text{Incorrect}_A}$, $\mathcal{D}_{J-}^{\text{Correct}_A}$, and $\mathcal{D}_{J-}^{\text{Incorrect}_A}$ displayed from left to right, showing the evaluation accuracy variations under different conditions.}
    \label{fig:4_dataset_heatmap}
\end{figure*}

\begin{table*}[t]
    \centering
    \small
    % \resizebox{\columnwidth}{!}{%
    \begin{tabular}{lcccc}
        \hline
        \textbf{Judge Model} & \textbf{GSM8K} & \textbf{MMLUPro} & \textbf{\makecell{MT-Bench Writing}} & \textbf{\makecell{Avg.}} \\
        \hline
        Mistral Large 2 & 5.67\% & 22.12\% & 33.00\% & 21.77\% \\
        Mistral Medium 3 & 5.33\% & 21.90\% & 29.66\% & 21.35\% \\
        Llama 3.1 405B & 3.67\% & 20.02\% & 28.33\% & 19.52\% \\
        Gemini 2.0 Flash Lite & 4.33\% & 16.81\% & 36.00\% & 17.23\% \\
        Mistral Small 3.1 & 5.33\% & 16.90\% & 30.00\% & 17.00\% \\
        Gemma 3 4B & 1.00\% & 17.21\% & 29.00\% & 16.93\% \\
        Gemma 3 27B & 4.00\% & 16.31\% & 32.66\% & 16.56\% \\
        Gemini 1.5 Flash 8B & 7.00\% & 15.69\% & 27.66\% & 15.90\% \\
        Gemma 3 12B & 5.00\% & 15.59\% & 27.00\% & 15.64\% \\
        Gemini 2.0 Flash & 2.67\% & 11.47\% & 28.33\% & 11.97\% \\
        Gemini 1.5 Flash & 4.00\% & 11.09\% & 23.00\% & 11.39\% \\
        \hline
    \end{tabular}%
    % }
    \caption{Model overconfidence across datasets, showing the difference between percentage of samples predicted as correct and percentage actually correct. Higher values indicate greater bias toward predicting correctness. The average (rightmost column) is computed as a weighted mean across all datasets based on sample count.}
    \label{tab:Prediction Proportions}
\end{table*}

% \subsection{Dataset-Level Finer-Grained Observations}
\subsection{In-Depth Dataset-Level Analysis}
\label{sec:4_2}
To address these questions and gain a more comprehensive understanding of these relationships, we conduct more fine-grained analyses beyond the dataset-level observations. Specifically, we investigate \emph{whether the correctness of responses generated by the agent model influences the judge model's behavior}. We split $\mathcal{D}_J$ using more fine-grained criteria based on two factors:
a) whether the judge model answers the question correctly, and
b) whether the agent model answers the question correctly.
Following this approach, we separate $\mathcal{D}_J$ into four subsets: 

\begin{itemize}
    \item $\mathcal{D}_{J+}^{\text{Correct}_A}$: Questions where both the judge model and the agent model answer correctly.
    \item $\mathcal{D}_{J+}^{\text{Incorrect}_A}$: Questions where the judge model answers correctly, but the agent model does not.
    \item $\mathcal{D}_{J-}^{\text{Correct}_A}$: Questions where the judge model answers incorrectly, but the agent model answers correctly.
    \item $\mathcal{D}_{J-}^{\text{Incorrect}_A}$: Questions where both the judge model and the agent model answer incorrectly.
\end{itemize}

We report the performance breakdown across the four subsets in Figure~\ref{fig:4_dataset_heatmap}. The subsets $\mathcal{D}_{J+}^{\text{Correct}_A}$ and $\mathcal{D}_{J-}^{\text{Correct}_A}$ consistently achieve higher scores, indicating that most LLMs perform better when the label is \texttt{Correct}. This suggests that evaluation outcomes are influenced more by label distribution than by the judge model's capability in the answer generation task. Revisiting Table~\ref{tab:full_analysis2}, the consistent superiority of $\mathcal{D}_{J+}$ over $\mathcal{D}_{J-}$ arises because the strong performance of $\mathcal{D}_{J+}^{\text{Correct}_A}$ outweighs the weaker results of $\mathcal{D}_{J+}^{\text{Incorrect}_A}$, yielding a higher overall score than the $\mathcal{D}_{J-}$ subsets.

\begin{figure*}[t]
\centering
\begin{tcolorbox}[width=\textwidth]
\small
\textbf{[System]} \\
Please act as an impartial judge and evaluate the quality of the response provided by an AI assistant to the user question displayed below. You will be given a reference answer and the assistant's answer. Your evaluation should focus solely on the correctness of the assistant's final answer. Begin by independently solving the user question step-by-step to verify the correctness of the response, and compare your final answer with both the reference answer and the assistant's final answer. Provide a brief explanation of your judgment, highlighting any differences and their significance. Be as objective as possible. After providing your explanation, state your final verdict by strictly following this format: "[[Correct]]" if the response's final answer is correct and "[[Incorrect]]" if it is not.

\vspace{0.5em}
\textbf{[User Question]} \\
Leo starts with 20 apples. He gives half to his sister. Then, he buys a new bag of 12 apples. After that, he uses 5 apples to bake a pie. How many apples does Leo have left?

\vspace{0.5em}
\textbf{[The Start of Reference Answer]} \\
Let's think step by step.
Leo begins with 20 apples.
...
The answer is 17.

\textbf{[The End of Reference Answer]}

\vspace{0.5em}
\textbf{[The Start of Assistant’s Answer]} \\
Let's think step by step.
Leo starts with 20 apples.
...
The answer is 11.

\textbf{[The End of Assistant’s Answer]}
\end{tcolorbox}

\caption{Example of Self-Reference-Guided Judgment on a Math Problem}
\label{fig:math_example_ref_guided_prompt}
\end{figure*}

To better understand the correlation between answer generation and judgment capabilities, we focus our analysis on the MMLU Pro and MT-Bench Writing datasets, temporarily excluding GSM8K-related datasets. This is because advanced models such as LLaMA 3.1 405B and Gemini 2.0 Flash achieve very high accuracy on GSM8K, resulting in too few instances in subsets like $\mathcal{D}_{J-}^{\text{Correct}A}$ and $\mathcal{D}_{J-}^{\text{Incorrect}_A}$ for meaningful evaluation. In the MMLU Pro dataset, it is evident that even high-performing LLMs like LLaMA 3.1 405B and Mistral Medium 3 achieve only around 60\% F1 score on $\mathcal{D}_{J+}^{\text{Incorrect}_A}$. Only a select few models, such as Gemini 2.0 Flash and Gemini 1.5 Flash, demonstrate strong performance in this subset. A similar trend is observed in the MT-Bench Writing subset, where model performance on $\mathcal{D}_{J+}^{\text{Incorrect}_A}$ remains low across the board, with the best models reaching an F1 score of only 0.44.

In Figure~\ref{fig:4_dataset_heatmap}, we observe that within the MMLU Pro dataset, Gemini 2.0 Flash Lite outperforms Gemini 2.0 Flash in the subset $\mathcal{D}_{J-}^{\text{Correct}A}$. To investigate this counterintuitive result, we analyze the distribution of ground-truth labels and model predictions across various judges and datasets. As shown in Table~\ref{tab:Prediction Proportions}, most LLMs exhibit a strong bias toward predicting \textit{Correct}. Since all ground-truth labels in $\mathcal{D}_{J-}^{\text{Correct}_A}$ are \textit{Correct}, weaker models may sometimes outperform stronger models from the same series due to this prediction bias.

% \subsection{Instance level observation}
% 在上一個section我們知道了response正確性會對correlation有很大的影響，所以我們算correlation的時候希望能移除agent response 對 correlation 的影響， 
% we measure the patial correlation of answer generation and answer judgment using Equation~\ref{eq:partial_corr}
% The results, shown in Figure~\ref{fig:capability_comparison}, reveal.... (先省略，我之後寫)

\subsection{Instance-Level Analysis}
While the previous analyses focused on dataset-level patterns, we now turn to instance-level analysis to obtain more fine-grained insights. Since response correctness strongly influences the correlation between generation and judgment abilities, we control for this confounding factor by measuring the partial correlation between answer generation and answer judgment using Equation~\ref{eq:partial_corr}, with results summarized in Table~\ref{tab:partial_correlation}. In most cases (25 out of 33), the correlations fall below 0.3, indicating only a weak association between the two abilities. A smaller number of cases (8 out of 33) show moderate alignment, with correlations between 0.3 and 0.5. Importantly, none of the cases exhibit strong correlation (above 0.5), which would suggest that generation and judgment rely on the same underlying mechanism. These findings reinforce our earlier observation that the correctness of the response being judged substantially influences evaluation outcomes, while also suggesting that an LLM's judgment ability is largely independent of its generation ability. 
This instance-level analysis complements the dataset-level observations: whereas aggregate results (Figure~\ref{fig:capability_comparison}) suggested a positive correlation between judgment and generation, the weak partial correlations at the instance level reveal that the two capabilities operate more independently than the dataset-level trends imply.

% The results, shown in Table~\ref{tab:partial_correlation}, reveal the partial correlation between answer judgment ability and answer generation ability across various LLMs. Our findings indicate that in most cases (25 out of 33), the correlations are lower than 0.3, which suggests only a weak association between these two abilities. Only a few instances (8 out of 33) demonstrate moderate semantic alignment with correlations in the range of 0.3 - 0.5. Notably, none of the analyzed cases exhibited strong correlation (above 0.5), which would have indicated that understanding and evaluation stem from the same source. This pattern further reinforces our earlier observation that the response being judged significantly influences judgment outcomes, while suggesting that an LLM's ability to judge answers is largely independent from its ability to generate answers. 
% This instance-level analysis adds an important dimension to our dataset-level observations: while dataset-level trends showed a positive correlation between answer judgment and generation abilities (as shown in Figure\ref{fig:capability_comparison}), the weak partial correlations at the instance level reveal that these two cognitive capabilities operate more independently than the aggregate data initially suggested. 
% Our quantitative findings provide concrete evidence that the relationship between answer generation and judgment abilities is predominantly weak (correlation < 0.3) in the vast majority of cases, helping to reconcile conflicting conclusions from previous research regarding these capabilities.

\begin{table}[]
    \centering
    \resizebox{\columnwidth}{!}{%
    \begin{tabular}{lccc}
        \toprule
        \textbf{Judge Model} & \textbf{GSM8K} & \textbf{MMLUPro} & \textbf{\makecell{MT-Bench\\Writing}} \\
        \midrule
        Llama 3.1 405B & \cellcolor{red!20}{0.1869} & \cellcolor{red!20}{0.2808} & \cellcolor{blue!20}{0.4448} \\
        Gemini 2.0 Flash & \cellcolor{red!20}{0.0864} & \cellcolor{red!20}{0.2862} & \cellcolor{blue!20}{0.3053} \\
        Gemini 2.0 Flash Lite & \cellcolor{blue!20}{0.3246} & \cellcolor{red!20}{0.2580} & \cellcolor{red!20}{0.1932} \\
        Gemini 1.5 Flash & \cellcolor{red!20}{0.1446} & \cellcolor{red!20}{0.2653} & \cellcolor{red!20}{0.2786} \\
        Gemini 1.5 Flash 8B & \cellcolor{blue!20}{0.3789} & \cellcolor{red!20}{0.1866} & \cellcolor{red!20}{0.1177} \\
        Gemma 3 4B & \cellcolor{blue!20}{0.3495} & \cellcolor{red!20}{0.1266} & \cellcolor{red!20}{0.1931} \\
        Gemma 3 12B & \cellcolor{blue!20}{0.3198} & \cellcolor{red!20}{0.1900} & \cellcolor{blue!20}{0.3365} \\
        Gemma 3 27B & \cellcolor{red!20}{0.0864} & \cellcolor{red!20}{0.2406} & \cellcolor{red!20}{0.1960} \\
        Mistral Small 3.1 & \cellcolor{red!20}{0.0828} & \cellcolor{red!20}{0.2207} & \cellcolor{red!20}{0.2240} \\
        Mistral Medium 3 & \cellcolor{red!20}{0.2800} & \cellcolor{red!20}{0.2729} & \cellcolor{blue!20}{0.4615} \\
        Mistral Large 2 & \cellcolor{red!20}{0.0628} & \cellcolor{red!20}{0.2443} & \cellcolor{red!20}{0.2754} \\
        \bottomrule
    \end{tabular}%
    }
    \caption{Partial correlation between answer generation and judgment capabilities across models and datasets. Weak and moderate correlations are highlighted with \colorbox{red!30}{red} and \colorbox{blue!30}{purple} backgrounds, respectively.}
    \label{tab:partial_correlation}
\end{table}

\begin{table*}[t]
    \centering
    \small
    % \resizebox{\columnwidth}{!}{%
    \begin{tabular}{lccc}
        \toprule
        \textbf{Judge Model} & \textbf{GSM8K} & \textbf{MMLU Pro} & \textbf{\makecell{MT-Bench\\Writing}} \\
        \midrule
        Llama 3.1 405B & \cellcolor{blue!20}{0.3950} (\textcolor{blue}{+0.2081}↑) & \cellcolor{green!20}{0.5719} (\textcolor{blue}{+0.2911}↑) & \cellcolor{green!20}{0.9283} (\textcolor{blue}{+0.4835}↑) \\
        Gemini 2.0 Flash & \cellcolor{blue!20}{0.4543} (\textcolor{blue}{+0.3679}↑) & \cellcolor{green!20}{0.5177} (\textcolor{blue}{+0.2315}↑) & \cellcolor{green!20}{0.8719} (\textcolor{blue}{+0.5666}↑) \\
        Gemini 2.0 Flash Lite & \cellcolor{blue!20}{0.3987} (\textcolor{blue}{+0.0741}↑) & \cellcolor{green!20}{0.5114} (\textcolor{blue}{+0.2534}↑) & \cellcolor{blue!20}{0.4202} (\textcolor{blue}{+0.2270}↑) \\
        Gemini 1.5 Flash & \cellcolor{blue!20}{0.4747} (\textcolor{blue}{+0.3301}↑) & \cellcolor{green!20}{0.5687} (\textcolor{blue}{+0.3034}↑) & \cellcolor{green!20}{0.8414} (\textcolor{blue}{+0.5628}↑) \\
        Gemini 1.5 Flash 8B & \cellcolor{green!20}{0.6916} (\textcolor{blue}{+0.3127}↑) & \cellcolor{blue!20}{0.4637} (\textcolor{blue}{+0.2771}↑) & \cellcolor{green!20}{0.6983} (\textcolor{blue}{+0.5806}↑) \\
        Gemma 3 4B & \cellcolor{green!20}{0.6243} (\textcolor{blue}{+0.2748}↑) & \cellcolor{blue!20}{0.4135} (\textcolor{blue}{+0.2869}↑) & \cellcolor{green!20}{0.8006} (\textcolor{blue}{+0.6075}↑) \\
        Gemma 3 12B & \cellcolor{blue!20}{0.4795} (\textcolor{blue}{+0.1597}↑) & \cellcolor{green!20}{0.5398} (\textcolor{blue}{+0.3498}↑) & \cellcolor{green!20}{0.6697} (\textcolor{blue}{+0.3332}↑) \\
        Gemma 3 27B & \cellcolor{red!20}{0.2585} (\textcolor{blue}{+0.1721}↑) & \cellcolor{green!20}{0.5620} (\textcolor{blue}{+0.3214}↑) & \cellcolor{green!20}{0.9036} (\textcolor{blue}{+0.7076}↑) \\
        Mistral Small 3.1 & \cellcolor{blue!20}{0.3140} (\textcolor{blue}{+0.2312}↑) & \cellcolor{green!20}{0.5430} (\textcolor{blue}{+0.3223}↑) & \cellcolor{green!20}{0.8688} (\textcolor{blue}{+0.6448}↑) \\
        Mistral Medium 3 & \cellcolor{blue!20}{0.4483} (\textcolor{blue}{+0.1683}↑) & \cellcolor{green!20}{0.5207} (\textcolor{blue}{+0.2478}↑) & \cellcolor{green!20}{0.9071} (\textcolor{blue}{+0.4456}↑) \\
        Mistral Large 2 & \cellcolor{green!20}{0.5931} (\textcolor{blue}{+0.5303}↑) & \cellcolor{green!20}{0.5869} (\textcolor{blue}{+0.3426}↑) & \cellcolor{green!20}{0.8079 (\textcolor{blue}{+0.5325}↑)} \\
        \bottomrule
    \end{tabular}%
    % }
    \caption{Partial correlation after applying the \textit{self-reference-guided} judgment method. Improvements over the CoT baseline are shown in \textcolor{blue}{blue} with upward arrows. Weak, moderate, and strong correlations are highlighted with \colorbox{red!30}{red}, \colorbox{blue!30}{purple}, and \colorbox{green!30}{green} backgrounds, respectively.}
    \label{tab:self_ref_partial_correlation}
\end{table*}

\section{Self-Reference-Guided Evaluation}
\label{sec:5}

\subsection{Goal and Methodology}
To strengthen the correlation between answer generation and judgment capabilities, we propose a \textit{self-reference-guided evaluation} approach as a replacement for the standard CoT method. Unlike traditional reference-guided methods~\cite{badshah2024referenceguidedverdictllmsasjudgesautomatic}, which rely on responses from stronger models or gold-standard answers as references, our approach leverages the judge model's own generated response as the reference during evaluation. Specifically, we use the answer generated by the judge model in Equation~\ref{judge-answer-generation}. Figure~\ref{fig:math_example_ref_guided_prompt} illustrates this method with a mathematical reasoning example. This design raises the following questions:

\begin{itemize}
\item Does using self-generated responses as references improve the correlation between answer generation and judgment capabilities compared to standard CoT?
\item How does the evaluation performance of self-reference-guided evaluation compare with traditional CoT?
\end{itemize}

\subsection{Results and Observations}

\paragraph{Correlation Enhancement.}
Table~\ref{tab:self_ref_partial_correlation} shows that our proposed method substantially improves the correlation compared to the standard CoT baseline in Table~\ref{tab:partial_correlation}. In 22 out of 33 cases, the correlations exceed 0.5, indicating strong alignment between answer generation and judgment capabilities. Another 10 cases fall into the moderate range, with only 1 case remaining weak. On average, the correlation increases by about 0.35, underscoring the effectiveness of the self-reference-guided approach in reducing the previously observed decoupling between the two abilities.

% As illustrated in Table \ref{tab:self_ref_partial_correlation}, we observe a substantial improvement in correlation strength compared to the standard CoT method. In most cases (22 out of 33), the correlations now exceed 0.5, indicating strong correlation between answer generation and judgment capabilities. On average, the correlation increased by approximately 0.35, demonstrating the effectiveness of the self-reference-guided approach in addressing the previously observed capability decoupling phenomenon. 

\begin{figure}[htbp]
    \centering
    \includegraphics[width=0.7\columnwidth]{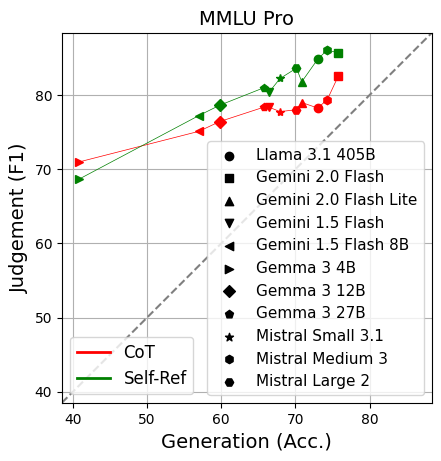} % 圖片文件的路徑
    \caption{Performance comparison between CoT and self-reference-guided evaluation methods on MMLU Pro, plotting answer generation accuracy against judgment F1 score for each model.}
    \label{fig:CoT vs Ref}
\end{figure}

\paragraph{Model-Specific Effects.}
Figure~\ref{fig:CoT vs Ref} shows the comparison between our self-reference-guided method and the standard CoT baseline in answer judgment performance. On the MMLU Pro dataset, the self-reference-guided method outperforms CoT once the judge model’s own answer generation accuracy exceeds 50\%. We emphasize that this specific threshold is not a universal rule and is expected to vary across different datasets. This may be related to the quality of the provided reference.

% We observe that when the accuracy of the judge model itself in the answer generation phase exceeds 50\%, the self-reference-guided method outperforms the CoT method. Conversely, when the accuracy in the answer generation phase is below 50\%, the CoT method outperforms the self-reference-guided method. This may be related to the quality of the provided reference.

% The experimental results suggest that the choice of evaluation strategies significantly influences the analytical outcomes. For instance, employing the reference-guided method reveals a substantial correlation between the LLMs capabilities in answer generation and answer judgment. In contrast, when utilizing the CoT approach, only the most sophisticated LLMs demonstrate minimal correlation between these two capabilities.

\subsection{Discussion and Practical Takeaway}
\label{sec:6}

Our experiments with CoT prompting show that answer generation and answer judgment are only weakly correlated. Consequently, selecting the top-performing model on a generation benchmark as a judge does not guarantee reliable evaluations. In contrast, our self-reference-guided strategy strengthens the connection between these abilities by using the model’s own answer as the reference, making generation accuracy a dependable proxy for judgment quality. This approach is particularly valuable when high-quality external references are unavailable, such as when gold labels are costly to obtain or access to stronger models is impractical. By leveraging self-generated outputs, our method aligns generation and judgment skills without external dependencies, offering a practical and robust path to reliable evaluation.

\section{Conclusion}
In this paper, we conducted a systematic examination of the correlation between performance in answer generation and answer judgment tasks using LLMs with standard CoT prompting. We evaluated 11 widely used LLMs on 21 benchmark subtasks. Our results show that, for most models, performance on answer generation is only weakly correlated with the ability to judge answers, indicating that strong generators are not necessarily reliable judges.
In addition, we found that \textit{self-reference-guided} evaluation methods can strengthen the correlation between generation and judgment capabilities. 
Based on these insights, we offer practical recommendations for selecting models to serve as judges, especially when external references like golden answers or outputs from stronger models are unavailable. In these situations, our work provides an accessible approach to judge model selection by using generation performance as a reliable proxy for judgment capability.
% Based on these insights, we offer practical recommendations for selecting models to serve as judges. Our work provides a more accessible approach to judge model selection by using generation performance as a proxy. 
This also mitigates the risk of generation and evaluation capabilities diverging as LLMs continue to develop.

\section{Limitations}
This study confronts inherent limitations due to the rapid evolution of LLMs and the specific nature of the evaluation tasks employed:

\paragraph{Evolving Landscape of LLMs.} The field of large language models is rapidly evolving, with new models continually introducing architectural and methodological advancements. Given this dynamic landscape, our study is necessarily constrained to the specific set of LLMs we evaluated. While our findings provide valuable insights into the current generation of models, future advancements may lead to changes in the relationship between answer generation and judgment capabilities, potentially strengthening or reshaping our observations. As such, ongoing research will be essential to understand how these relationships evolve as models continue to improve.
    
\paragraph{Self-Reference-Guided Evaluation Scope.} The self-reference-guided method has shown potential in enhancing the correlation between answer generation and answer judgment tasks. However, this study has employed the self-reference-guided method specifically within the context of pointwise answer judgment tasks, where the references needed are clearly defined as the judge model's responses to questions. The applicability of this method to more complex evaluation formats and tasks, such as pairwise comparison or listwise ranking, remains uncertain. Further research is required to determine how references can be effectively generated and utilized in diverse evaluative contexts beyond simple pointwise answer judgment.

\paragraph{Lack of Multi-Turn Interaction Analysis.} Our study focuses exclusively on single-turn interactions, a simplification of real-world applications where judgments often occur in multi-turn contexts. In interactive settings, models must adapt their generation and evaluation to prior context, which we do not address here. We intentionally restrict our scope to single-turn interactions because they underlie most LLM-as-Judge benchmarks and offer a clearer lens for our research question. Nevertheless, this constitutes a limitation. Future work should extend our analysis to multi-turn dialogues to assess whether the observed correlation patterns and the effectiveness of self-reference-guided evaluation hold in more complex interactive scenarios.

% Our study exclusively examines single-turn interactions, which simplifies real-world LLM applications where judgments often occur in multi-turn contexts. In interactive scenarios, models must adjust their generation and evaluation based on previous exchanges - a dynamic capability our research does not address. While we deliberately focused on single-turn interactions as they form the foundation of most LLM-as-Judge benchmarks and provide clarity for our research question, this represents a limitation. Future work should extend our findings to multi-turn dialogues to determine whether the observed correlation patterns and effectiveness of self-reference-guided evaluation persist in more complex interactive settings.

\paragraph{Potential for Error Propagation} A key consideration for the self-reference-guided method is the potential for error propagation. Since the approach uses the judge's own output as the reference, an incorrect reference answer can lead to flawed evaluations. For instance, a judge model might incorrectly penalize an agent's correct response simply because it fails to match its own erroneous reference. This risk is empirically supported by our own findings, which demonstrate that the effectiveness of the self-reference-guided method is directly tied to the judge model's generation accuracy. While this is a limitation, it also reinforces our primary recommendation to apply this method using judge models with high generation performance for the target domain.
These limitations underscore the necessity for ongoing research to continually reassess and validate the applicability of our findings as the technology evolves and to expand the methodological framework to include a wider variety of evaluation tasks in different contexts.

\section*{Acknowledgments}
This work was supported by National Science and
Technology Council, Taiwan, under grants NSTC
113-2634-F-002-003- and 114-2221-E-002-070-
MY3, and Ministry of Education (MOE), Taiwan,
under grant NTU-114L900901.

\section*{Use of AI Assistants}
We sincerely appreciate the assistance provided by ChatGPT in refining our manuscript. 
ChatGPT offered suggestions for improving the clarity and conciseness of our writing, helped restructure key sections for better readability, and contributed to refining our research terminology. 
% Additionally, ChatGPT provided constructive feedback on our experimental framework and helped streamline our recommendations for selecting judge models. 
While the final content remains our own, these contributions enhanced the presentation of our work.

\bibliography{custom}

\begin{thebibliography}{39}
\providecommand{\natexlab}[1]{#1}

\bibitem[{Badshah and Sajjad(2024)}]{badshah2024referenceguidedverdictllmsasjudgesautomatic}
Sher Badshah and Hassan Sajjad. 2024.
\newblock \href {https://arxiv.org/abs/2408.09235} {Reference-guided verdict: Llms-as-judges in automatic evaluation of free-form text}.
\newblock \emph{Preprint}, arXiv:2408.09235.

\bibitem[{Bi et~al.(2024)Bi, Zhang, Xue, Ou, Ji, Zheng, and Chen}]{bi-etal-2024-oceangpt}
Zhen Bi, Ningyu Zhang, Yida Xue, Yixin Ou, Daxiong Ji, Guozhou Zheng, and Huajun Chen. 2024.
\newblock \href {https://doi.org/10.18653/v1/2024.acl-long.184} {{O}cean{GPT}: A large language model for ocean science tasks}.
\newblock In \emph{Proceedings of the 62nd Annual Meeting of the Association for Computational Linguistics (Volume 1: Long Papers)}, pages 3357--3372, Bangkok, Thailand. Association for Computational Linguistics.

\bibitem[{Chan et~al.(2024)Chan, Chen, Su, Yu, Xue, Zhang, Fu, and Liu}]{chan2024chateval}
Chi-Min Chan, Weize Chen, Yusheng Su, Jianxuan Yu, Wei Xue, Shanghang Zhang, Jie Fu, and Zhiyuan Liu. 2024.
\newblock \href {https://openreview.net/forum?id=FQepisCUWu} {Chateval: Towards better {LLM}-based evaluators through multi-agent debate}.
\newblock In \emph{The Twelfth International Conference on Learning Representations}.

\bibitem[{Chen et~al.(2024{\natexlab{a}})Chen, Chen, Liu, Jiang, and Wang}]{chen-etal-2024-humans}
Guiming~Hardy Chen, Shunian Chen, Ziche Liu, Feng Jiang, and Benyou Wang. 2024{\natexlab{a}}.
\newblock \href {https://doi.org/10.18653/v1/2024.emnlp-main.474} {Humans or {LLM}s as the judge? a study on judgement bias}.
\newblock In \emph{Proceedings of the 2024 Conference on Empirical Methods in Natural Language Processing}, pages 8301--8327, Miami, Florida, USA. Association for Computational Linguistics.

\bibitem[{Chen et~al.(2024{\natexlab{b}})Chen, Saha, and Bansal}]{chen-etal-2024-reconcile}
Justin Chen, Swarnadeep Saha, and Mohit Bansal. 2024{\natexlab{b}}.
\newblock \href {https://doi.org/10.18653/v1/2024.acl-long.381} {{R}e{C}oncile: Round-table conference improves reasoning via consensus among diverse {LLM}s}.
\newblock In \emph{Proceedings of the 62nd Annual Meeting of the Association for Computational Linguistics (Volume 1: Long Papers)}, pages 7066--7085, Bangkok, Thailand. Association for Computational Linguistics.

\bibitem[{Chiang et~al.(2024)Chiang, Zheng, Sheng, Angelopoulos, Li, Li, Zhu, Zhang, Jordan, Gonzalez, and Stoica}]{pmlr-v235-chiang24b}
Wei-Lin Chiang, Lianmin Zheng, Ying Sheng, Anastasios~Nikolas Angelopoulos, Tianle Li, Dacheng Li, Banghua Zhu, Hao Zhang, Michael Jordan, Joseph~E. Gonzalez, and Ion Stoica. 2024.
\newblock \href {https://proceedings.mlr.press/v235/chiang24b.html} {Chatbot arena: An open platform for evaluating {LLM}s by human preference}.
\newblock In \emph{Proceedings of the 41st International Conference on Machine Learning}, volume 235 of \emph{Proceedings of Machine Learning Research}, pages 8359--8388. PMLR.

\bibitem[{Cobbe et~al.(2021)Cobbe, Kosaraju, Bavarian, Chen, Jun, Kaiser, Plappert, Tworek, Hilton, Nakano, Hesse, and Schulman}]{cobbe_training_2021}
Karl Cobbe, Vineet Kosaraju, Mohammad Bavarian, Mark Chen, Heewoo Jun, Lukasz Kaiser, Matthias Plappert, Jerry Tworek, Jacob Hilton, Reiichiro Nakano, Christopher Hesse, and John Schulman. 2021.
\newblock \href {https://doi.org/10.48550/arXiv.2110.14168} {Training {Verifiers} to {Solve} {Math} {Word} {Problems}}.
\newblock \emph{arXiv preprint}.
\newblock ArXiv:2110.14168.

\bibitem[{Dubey et~al.(2024)Dubey, Jauhri, Pandey, Kadian, Al-Dahle, Letman, Mathur, Schelten, Yang, Fan et~al.}]{dubey2024llama}
Abhimanyu Dubey, Abhinav Jauhri, Abhinav Pandey, Abhishek Kadian, Ahmad Al-Dahle, Aiesha Letman, Akhil Mathur, Alan Schelten, Amy Yang, Angela Fan, et~al. 2024.
\newblock The llama 3 herd of models.
\newblock \emph{arXiv preprint arXiv:2407.21783}.

\bibitem[{Fei et~al.(2024)Fei, Shen, Zhu, Zhou, Han, Huang, Zhang, Chen, Yin, Shen, Ge, and Ng}]{fei-etal-2024-lawbench}
Zhiwei Fei, Xiaoyu Shen, Dawei Zhu, Fengzhe Zhou, Zhuo Han, Alan Huang, Songyang Zhang, Kai Chen, Zhixin Yin, Zongwen Shen, Jidong Ge, and Vincent Ng. 2024.
\newblock \href {https://doi.org/10.18653/v1/2024.emnlp-main.452} {{L}aw{B}ench: Benchmarking legal knowledge of large language models}.
\newblock In \emph{Proceedings of the 2024 Conference on Empirical Methods in Natural Language Processing}, pages 7933--7962, Miami, Florida, USA. Association for Computational Linguistics.

\bibitem[{Gemini~Team(2023)}]{Anil2023GeminiAF}
Google Gemini~Team. 2023.
\newblock \href {https://arxiv.org/pdf/2312.11805} {Gemini: A family of highly capable multimodal models}.
\newblock \emph{ArXiv}, abs/2312.11805.

\bibitem[{Gemini~Team(2024)}]{gemini15}
Google Gemini~Team. 2024.
\newblock \href {https://storage.googleapis.com/deepmind-media/gemini/gemini_v1_5_report.pdf} {Gemini 1.5: Unlocking multimodal understanding across millions of tokens of context}.

\bibitem[{{Gemma Team}(2025)}]{gemmateam2025gemma3technicalreport}
{Gemma Team}. 2025.
\newblock \href {https://arxiv.org/abs/2503.19786} {Gemma 3 technical report}.
\newblock \emph{Preprint}, arXiv:2503.19786.

\bibitem[{Hendrycks et~al.(2021{\natexlab{a}})Hendrycks, Burns, Basart, Zou, Mazeika, Song, and Steinhardt}]{hendrycks2021measuring}
Dan Hendrycks, Collin Burns, Steven Basart, Andy Zou, Mantas Mazeika, Dawn Song, and Jacob Steinhardt. 2021{\natexlab{a}}.
\newblock \href {https://openreview.net/forum?id=d7KBjmI3GmQ} {Measuring massive multitask language understanding}.
\newblock In \emph{International Conference on Learning Representations}.

\bibitem[{Hendrycks et~al.(2021{\natexlab{b}})Hendrycks, Burns, Kadavath, Arora, Basart, Tang, Song, and Steinhardt}]{hendrycksmath2021}
Dan Hendrycks, Collin Burns, Saurav Kadavath, Akul Arora, Steven Basart, Eric Tang, Dawn Song, and Jacob Steinhardt. 2021{\natexlab{b}}.
\newblock Measuring mathematical problem solving with the math dataset.
\newblock \emph{NeurIPS}.

\bibitem[{Kojima et~al.(2022)Kojima, Gu, Reid, Matsuo, and Iwasawa}]{NEURIPS2022_8bb0d291}
Takeshi Kojima, Shixiang~(Shane) Gu, Machel Reid, Yutaka Matsuo, and Yusuke Iwasawa. 2022.
\newblock \href {https://proceedings.neurips.cc/paper_files/paper/2022/file/8bb0d291acd4acf06ef112099c16f326-Paper-Conference.pdf} {Large language models are zero-shot reasoners}.
\newblock In \emph{Advances in Neural Information Processing Systems}, volume~35, pages 22199--22213. Curran Associates, Inc.

\bibitem[{Koo et~al.(2024)Koo, Lee, Raheja, Park, Kim, and Kang}]{koo-etal-2024-benchmarking}
Ryan Koo, Minhwa Lee, Vipul Raheja, Jong~Inn Park, Zae~Myung Kim, and Dongyeop Kang. 2024.
\newblock \href {https://doi.org/10.18653/v1/2024.findings-acl.29} {Benchmarking cognitive biases in large language models as evaluators}.
\newblock In \emph{Findings of the Association for Computational Linguistics: ACL 2024}, pages 517--545, Bangkok, Thailand. Association for Computational Linguistics.

\bibitem[{Lee et~al.(2024)Lee, Kim, Yousefpour, Seo, Yoo, and Yu}]{lee-etal-2024-aligning}
Sangkyu Lee, Sungdong Kim, Ashkan Yousefpour, Minjoon Seo, Kang~Min Yoo, and Youngjae Yu. 2024.
\newblock \href {https://doi.org/10.18653/v1/2024.acl-long.617} {Aligning large language models by on-policy self-judgment}.
\newblock In \emph{Proceedings of the 62nd Annual Meeting of the Association for Computational Linguistics (Volume 1: Long Papers)}, pages 11442--11459, Bangkok, Thailand. Association for Computational Linguistics.

\bibitem[{Li et~al.(2025)Li, Jiang, Huang, Beigi, Zhao, Tan, Bhattacharjee, Jiang, Chen, Wu, Shu, Cheng, and Liu}]{li2025generationjudgmentopportunitieschallenges}
Dawei Li, Bohan Jiang, Liangjie Huang, Alimohammad Beigi, Chengshuai Zhao, Zhen Tan, Amrita Bhattacharjee, Yuxuan Jiang, Canyu Chen, Tianhao Wu, Kai Shu, Lu~Cheng, and Huan Liu. 2025.
\newblock \href {https://arxiv.org/abs/2411.16594} {From generation to judgment: Opportunities and challenges of llm-as-a-judge}.
\newblock \emph{Preprint}, arXiv:2411.16594.

\bibitem[{Li et~al.(2024{\natexlab{a}})Li, Dong, Chen, Su, Zhou, Ai, Ye, and Liu}]{li2024llmsasjudgescomprehensivesurveyllmbased}
Haitao Li, Qian Dong, Junjie Chen, Huixue Su, Yujia Zhou, Qingyao Ai, Ziyi Ye, and Yiqun Liu. 2024{\natexlab{a}}.
\newblock \href {https://arxiv.org/abs/2412.05579} {Llms-as-judges: A comprehensive survey on llm-based evaluation methods}.
\newblock \emph{Preprint}, arXiv:2412.05579.

\bibitem[{Li et~al.(2024{\natexlab{b}})Li, Chiang, Frick, Dunlap, Wu, Zhu, Gonzalez, and Stoica}]{li2024crowdsourceddatahighqualitybenchmarks}
Tianle Li, Wei-Lin Chiang, Evan Frick, Lisa Dunlap, Tianhao Wu, Banghua Zhu, Joseph~E. Gonzalez, and Ion Stoica. 2024{\natexlab{b}}.
\newblock \href {https://arxiv.org/abs/2406.11939} {From crowdsourced data to high-quality benchmarks: Arena-hard and benchbuilder pipeline}.
\newblock \emph{Preprint}, arXiv:2406.11939.

\bibitem[{Liang et~al.(2024)Liang, Ye, Han, Lai, Zhang, Huang, and Wei}]{liang-etal-2024-debatrix}
Jingcong Liang, Rong Ye, Meng Han, Ruofei Lai, Xinyu Zhang, Xuanjing Huang, and Zhongyu Wei. 2024.
\newblock \href {https://doi.org/10.18653/v1/2024.findings-acl.868} {Debatrix: Multi-dimensional debate judge with iterative chronological analysis based on {LLM}}.
\newblock In \emph{Findings of the Association for Computational Linguistics: ACL 2024}, pages 14575--14595, Bangkok, Thailand. Association for Computational Linguistics.

\bibitem[{Lin and Chen(2023)}]{lin-chen-2023-llm}
Yen-Ting Lin and Yun-Nung Chen. 2023.
\newblock \href {https://doi.org/10.18653/v1/2023.nlp4convai-1.5} {{LLM}-eval: Unified multi-dimensional automatic evaluation for open-domain conversations with large language models}.
\newblock In \emph{Proceedings of the 5th Workshop on NLP for Conversational AI (NLP4ConvAI 2023)}, pages 47--58, Toronto, Canada. Association for Computational Linguistics.

\bibitem[{Liu et~al.(2024)Liu, Shi, Fabbri, Zhao, Wang, Wu, Joty, and Cohan}]{liu_reife_2024}
Yixin Liu, Kejian Shi, Alexander~R. Fabbri, Yilun Zhao, Peifeng Wang, Chien-Sheng Wu, Shafiq Joty, and Arman Cohan. 2024.
\newblock \href {https://doi.org/10.48550/arXiv.2410.07069} {{ReIFE}: {Re}-evaluating {Instruction}-{Following} {Evaluation}}.
\newblock \emph{arXiv preprint}.
\newblock ArXiv:2410.07069 [cs].

\bibitem[{Mirzadeh et~al.(2025)Mirzadeh, Alizadeh, Shahrokhi, Tuzel, Bengio, and Farajtabar}]{mirzadeh2025gsmsymbolic}
Seyed~Iman Mirzadeh, Keivan Alizadeh, Hooman Shahrokhi, Oncel Tuzel, Samy Bengio, and Mehrdad Farajtabar. 2025.
\newblock \href {https://openreview.net/forum?id=AjXkRZIvjB} {{GSM}-symbolic: Understanding the limitations of mathematical reasoning in large language models}.
\newblock In \emph{The Thirteenth International Conference on Learning Representations}.

\bibitem[{{Mistral AI Team}(2024{\natexlab{a}})}]{ai_large_2024}
{Mistral AI Team}. 2024{\natexlab{a}}.
\newblock \href {https://mistral.ai/news/mistral-large-2407/} {Large {Enough}}.
\newblock Section: news.

\bibitem[{{Mistral AI Team}(2024{\natexlab{b}})}]{ai_ministral_2024}
{Mistral AI Team}. 2024{\natexlab{b}}.
\newblock \href {https://mistral.ai/news/ministraux/} {Un {Ministral}, des {Ministraux}}.
\newblock Section: news.

\bibitem[{Raina et~al.(2024)Raina, Liusie, and Gales}]{raina-etal-2024-llm}
Vyas Raina, Adian Liusie, and Mark Gales. 2024.
\newblock \href {https://doi.org/10.18653/v1/2024.emnlp-main.427} {Is {LLM}-as-a-judge robust? investigating universal adversarial attacks on zero-shot {LLM} assessment}.
\newblock In \emph{Proceedings of the 2024 Conference on Empirical Methods in Natural Language Processing}, pages 7499--7517, Miami, Florida, USA. Association for Computational Linguistics.

\bibitem[{Shi et~al.(2024)Shi, Ma, Liang, Ma, and Vosoughi}]{shi_judging_2024}
Lin Shi, Chiyu Ma, Wenhua Liang, Weicheng Ma, and Soroush Vosoughi. 2024.
\newblock \href {https://doi.org/10.48550/arXiv.2406.07791} {Judging the {Judges}: {A} {Systematic} {Study} of {Position} {Bias} in {LLM}-as-a-{Judge}}.
\newblock \emph{arXiv preprint}.
\newblock ArXiv:2406.07791 [cs].

\bibitem[{Tan et~al.(2025)Tan, Zhuang, Montgomery, Tang, Cuadron, Wang, Popa, and Stoica}]{tan2025judgebench}
Sijun Tan, Siyuan Zhuang, Kyle Montgomery, William~Yuan Tang, Alejandro Cuadron, Chenguang Wang, Raluca Popa, and Ion Stoica. 2025.
\newblock \href {https://openreview.net/forum?id=G0dksFayVq} {Judgebench: A benchmark for evaluating {LLM}-based judges}.
\newblock In \emph{The Thirteenth International Conference on Learning Representations}.

\bibitem[{Verga et~al.(2024)Verga, Hofstatter, Althammer, Su, Piktus, Arkhangorodsky, Xu, White, and Lewis}]{verga2024replacingjudgesjuriesevaluating}
Pat Verga, Sebastian Hofstatter, Sophia Althammer, Yixuan Su, Aleksandra Piktus, Arkady Arkhangorodsky, Minjie Xu, Naomi White, and Patrick Lewis. 2024.
\newblock \href {https://arxiv.org/abs/2404.18796} {Replacing judges with juries: Evaluating llm generations with a panel of diverse models}.
\newblock \emph{Preprint}, arXiv:2404.18796.

\bibitem[{Wang et~al.(2024{\natexlab{a}})Wang, Li, Chen, Cai, Zhu, Lin, Cao, Kong, Liu, Liu, and Sui}]{wang-etal-2024-large-language-models-fair}
Peiyi Wang, Lei Li, Liang Chen, Zefan Cai, Dawei Zhu, Binghuai Lin, Yunbo Cao, Lingpeng Kong, Qi~Liu, Tianyu Liu, and Zhifang Sui. 2024{\natexlab{a}}.
\newblock \href {https://doi.org/10.18653/v1/2024.acl-long.511} {Large language models are not fair evaluators}.
\newblock In \emph{Proceedings of the 62nd Annual Meeting of the Association for Computational Linguistics (Volume 1: Long Papers)}, pages 9440--9450, Bangkok, Thailand. Association for Computational Linguistics.

\bibitem[{Wang et~al.(2024{\natexlab{b}})Wang, Yu, Yao, Zeng, Yang, Wang, Chen, Jiang, Xie, Wang, Xie, Ye, Zhang, and Zhang}]{wang2024pandalm}
Yidong Wang, Zhuohao Yu, Wenjin Yao, Zhengran Zeng, Linyi Yang, Cunxiang Wang, Hao Chen, Chaoya Jiang, Rui Xie, Jindong Wang, Xing Xie, Wei Ye, Shikun Zhang, and Yue Zhang. 2024{\natexlab{b}}.
\newblock \href {https://openreview.net/forum?id=5Nn2BLV7SB} {Panda{LM}: An automatic evaluation benchmark for {LLM} instruction tuning optimization}.
\newblock In \emph{The Twelfth International Conference on Learning Representations}.

\bibitem[{Wang et~al.(2024{\natexlab{c}})Wang, Ma, Zhang, Ni, Chandra, Guo, Ren, Arulraj, He, Jiang, Li, KU, Wang, Zhuang, Fan, Yue, and Chen}]{NEURIPS2024_ad236edc}
Yubo Wang, Xueguang Ma, Ge~Zhang, Yuansheng Ni, Abhranil Chandra, Shiguang Guo, Weiming Ren, Aaran Arulraj, Xuan He, Ziyan Jiang, Tianle Li, Max KU, Kai Wang, Alex Zhuang, Rongqi Fan, Xiang Yue, and Wenhu Chen. 2024{\natexlab{c}}.
\newblock \href {https://proceedings.neurips.cc/paper_files/paper/2024/file/ad236edc564f3e3156e1b2feafb99a24-Paper-Datasets_and_Benchmarks_Track.pdf} {Mmlu-pro: A more robust and challenging multi-task language understanding benchmark}.
\newblock In \emph{Advances in Neural Information Processing Systems}, volume~37, pages 95266--95290. Curran Associates, Inc.

\bibitem[{Wei et~al.(2024{\natexlab{a}})Wei, He, Xia, Wong, Lin, and Han}]{wei2024systematicevaluationllmasajudgellm}
Hui Wei, Shenghua He, Tian Xia, Andy Wong, Jingyang Lin, and Mei Han. 2024{\natexlab{a}}.
\newblock \href {https://arxiv.org/abs/2408.13006} {Systematic evaluation of llm-as-a-judge in llm alignment tasks: Explainable metrics and diverse prompt templates}.
\newblock \emph{Preprint}, arXiv:2408.13006.

\bibitem[{Wei et~al.(2022)Wei, Wang, Schuurmans, Bosma, ichter, Xia, Chi, Le, and Zhou}]{NEURIPS2022_9d560961}
Jason Wei, Xuezhi Wang, Dale Schuurmans, Maarten Bosma, brian ichter, Fei Xia, Ed~Chi, Quoc~V Le, and Denny Zhou. 2022.
\newblock \href {https://proceedings.neurips.cc/paper_files/paper/2022/file/9d5609613524ecf4f15af0f7b31abca4-Paper-Conference.pdf} {Chain-of-thought prompting elicits reasoning in large language models}.
\newblock In \emph{Advances in Neural Information Processing Systems}, volume~35, pages 24824--24837. Curran Associates, Inc.

\bibitem[{Wei et~al.(2024{\natexlab{b}})Wei, Wu, Huang, and Chen}]{wei_unveiling_2024}
Sheng-Lun Wei, Cheng-Kuang Wu, Hen-Hsen Huang, and Hsin-Hsi Chen. 2024{\natexlab{b}}.
\newblock \href {https://doi.org/10.18653/v1/2024.findings-acl.333} {Unveiling selection biases: Exploring order and token sensitivity in large language models}.
\newblock In \emph{Findings of the Association for Computational Linguistics: ACL 2024}, pages 5598--5621, Bangkok, Thailand. Association for Computational Linguistics.

\bibitem[{Zeng et~al.(2025)Zeng, Chen, Liu, Jiang, and Jia}]{zeng2025mrgsmk}
Zhongshen Zeng, Pengguang Chen, Shu Liu, Haiyun Jiang, and Jiaya Jia. 2025.
\newblock \href {https://openreview.net/forum?id=br4H61LOoI} {{MR}-{GSM}8k: A meta-reasoning benchmark for large language model evaluation}.
\newblock In \emph{The Thirteenth International Conference on Learning Representations}.

\bibitem[{Zheng et~al.(2023)Zheng, Chiang, Sheng, Zhuang, Wu, Zhuang, Lin, Li, Li, Xing, Zhang, Gonzalez, and Stoica}]{NEURIPS2023_91f18a12}
Lianmin Zheng, Wei-Lin Chiang, Ying Sheng, Siyuan Zhuang, Zhanghao Wu, Yonghao Zhuang, Zi~Lin, Zhuohan Li, Dacheng Li, Eric Xing, Hao Zhang, Joseph~E Gonzalez, and Ion Stoica. 2023.
\newblock \href {https://proceedings.neurips.cc/paper_files/paper/2023/file/91f18a1287b398d378ef22505bf41832-Paper-Datasets_and_Benchmarks.pdf} {Judging llm-as-a-judge with mt-bench and chatbot arena}.
\newblock In \emph{Advances in Neural Information Processing Systems}, volume~36, pages 46595--46623. Curran Associates, Inc.

\bibitem[{Zhu et~al.(2023)Zhu, Wang, and Wang}]{zhu_judgelm_2023}
Lianghui Zhu, Xinggang Wang, and Xinlong Wang. 2023.
\newblock \href {https://doi.org/10.48550/arXiv.2310.17631} {{JudgeLM}: {Fine}-tuned {Large} {Language} {Models} are {Scalable} {Judges}}.
\newblock \emph{arXiv preprint}.
\newblock ArXiv:2310.17631 [cs].

\end{thebibliography}

\appendix
\clearpage

\section{Additional Model Information}
\label{appendix:model_info}
To provide a comprehensive overview of the models used in our experiments, we list the agent and judge models along with their corresponding API endpoints in Table~\ref{model_info}. For Llama 3.1 models, we leverage the API provided by SambaNova\footnote{\url{https://cloud.sambanova.ai}} to optimize the efficiency and scalability of our experiments. For the Mistral series, we utilize APIs provided by Mistral AI, while for the Gemini series and Gemma series, we use APIs provided by Google.
These models were selected based on their performance and availability, ensuring a diverse set of architectures for evaluation. All model inference was conducted with temperature set to 0 to ensure reproducibility of our experiments.

\begin{table}[htbp]
\centering
\scalebox{0.7}{
\begin{tabular}{ll}
\toprule
\textbf{Model (Size)} & \textbf{API Endpoint} \\
\midrule
\multicolumn{2}{c}{\textbf{Agent Models}} \\
\midrule
Llama 3.1 8B & \texttt{Meta-Llama-3.1-8B-Instruct} \\
Llama 3.1 70B & \texttt{Meta-Llama-3.1-70B-Instruct} \\
Ministral 8B & \texttt{ministral-8b-2410} \\
\midrule
\multicolumn{2}{c}{\textbf{Judge Models}} \\
\midrule
Llama 3.1 405B & \texttt{Meta-Llama-3.1-405B-Instruct} \\
Mistral Small 3.1 & \texttt{mistral-small-2503} \\
Mistral Medium 3 & \texttt{mistral-medium-2505} \\
Mistral Large 2 & \texttt{mistral-large-2411} \\
Gemma 3 4B & \texttt{gemma-3-4b-it} \\
Gemma 3 12B & \texttt{gemma-3-12b-it} \\
Gemma 3 27B & \texttt{gemma-3-27b-it} \\
Gemini 1.5 Flash 8B & \texttt{gemini-1.5-flash-8b-001} \\
Gemini 1.5 Flash & \texttt{gemini-1.5-flash-002} \\
Gemini 2.0 Flash Lite & \texttt{gemini-2.0-flash-lite-001} \\
Gemini 2.0 Flash & \texttt{gemini-2.0-flash-001} \\
\bottomrule
\end{tabular}
}
\caption{Model endpoints used in our experiments}
\label{model_info}
\end{table}

% \section{Prompt templates}
% We provide a comprehensive list of all prompt templates used in our experiments, including those for answer generation, CoT answer judgment, and reference-guided answer judgment. The answer generation prompt is derived from the original dataset's approach to CoT reasoning~\citep{NEURIPS2022_8bb0d291, NEURIPS2024_ad236edc}.
% The CoT answer judgment and reference-guided answer judgment prompts are adapted with slight modifications from \cite{NEURIPS2023_91f18a12}. These templates are presented in Figures \ref{fig:GSM8K_answer_generation_prompts} \ref{fig:MMLU_Pro_answer_generation_prompts} \ref{fig:pairwise_evaluation_prompt}, \ref{fig:cot_prompt} \ref{fig:meta_evaluation_prompt} and \ref{fig:ref_guided_prompt} \ref{fig:reference_guided_meta_evaluation_prompt}, corresponding to their respective tasks.

\section{Prompt Templates}
For reproducibility, we provide all prompt templates used in our experiments. The answer generation prompts, derived from the original datasets’ CoT reasoning approaches~\citep{NEURIPS2022_8bb0d291, NEURIPS2024_ad236edc, NEURIPS2023_91f18a12}, are shown in Figures~\ref{fig:GSM8K_answer_generation_prompts}, \ref{fig:MMLU_Pro_answer_generation_prompts}, and \ref{fig:chatbot_arena_answer_generation_prompt}. The evaluation templates, including both CoT answer judgment and self-reference-guided evaluation prompts, are adapted with slight modifications from \citet{NEURIPS2023_91f18a12} and are presented in Figures~\ref{fig:MMLU_cot_prompt}, \ref{fig:chatbot_arena_cot_prompt}, \ref{fig:MMLU_ref_guided_prompt}, and \ref{fig:chatbot_arena_ref_guided_prompt}, corresponding to their respective evaluation tasks.

\begin{figure*}[tp]
\centering
\begin{tcolorbox}[width=\textwidth]
\small
\textbf{[System]} \\
The question is: \{question\}. Let’s think step by step and then finish your answer with "The answer is (arabic numerals):"
\end{tcolorbox}
\caption{Prompt templates for CoT answer generation in the GSM8K, GSM-Symbolic-P1 and GSM-Symbolic-P2 datasets.}
\label{fig:GSM8K_answer_generation_prompts}
\end{figure*}

\begin{figure*}[tp]
\centering
\begin{tcolorbox}[width=\textwidth]
\small
\textbf{[System]} \\
The following are multiple-choice questions (with answers) about \{\textit{category}\}. Think step by step and then output the answer in the format of "The answer is (\textit{X})" at the end.

\vspace{0.5em}

\textbf{Question:} \{question\} \\
\textbf{Options:} \\
\textit{A. \{Option 1\}} \\
\textit{B. \{Option 2\}} \\
\textit{C. \{Option 3\}} \\
\textit{D. \{Option 4\}} \\
\vdots

\vspace{0.5em}

\textbf{Answer:} Let's think step by step.

\end{tcolorbox}
\caption{Prompt templates for CoT answer generation in the MMLU Pro datasets.}
\label{fig:MMLU_Pro_answer_generation_prompts}
\end{figure*}

\begin{figure*}[tp]
\centering
\begin{tcolorbox}[width=\textwidth]
\small
\textbf{[System]} \\
Please act as an impartial judge and evaluate the quality of the responses provided by two AI assistants to the user question displayed below. Your evaluation should consider correctness and helpfulness. You will be given assistant A's answer, and assistant B's answer. Your job is to evaluate which assistant's answer is better. You should independently solve the user question step-by-step first. Then compare both assistants' answers with your answer. Identify and correct any mistakes. Avoid any position biases and ensure that the order in which the responses were presented does not influence your decision. Do not allow the length of the responses to influence your evaluation. Do not favor certain names of the assistants. Be as objective as possible. After providing your explanation, output your final verdict by strictly following this format: "[[A]]" if assistant A is better, "[[B]]" if assistant B is better, and "[[C]]" for a tie.

\vspace{0.5em}
\textbf{[User Question]} \\
\{Question\} \\
\vspace{0.5em}
\textbf{[The Start of Assistant A's Answer]} \\
\{Answer A\} \\
\textbf{[The End of Assistant A's Answer]} \\
\vspace{0.5em}
\textbf{[The Start of Assistant B's Answer]} \\
\{Answer B\} \\
\textbf{[The End of Assistant B's Answer]}
\end{tcolorbox}
\caption{Prompt template for CoT answer generation in the Chatbot Arena and MT-Bench subtasks datasets.}
\label{fig:chatbot_arena_answer_generation_prompt}
\end{figure*}

% ---------------------------------------- CoT Prompt ----------------------------------------------

\begin{figure*}[tp]
\centering
\begin{tcolorbox}[width=\textwidth]
\small
\textbf{[System]} \\
Please act as an impartial judge and evaluate the quality of the response provided by an AI assistant to the user question displayed below. Your evaluation should focus solely on the correctness of the assistant's final answer. Begin by independently solving the user question step-by-step to verify the correctness of the response. Compare your final answer to the assistant's final answer. Provide a brief explanation of your judgment. Be as objective as possible. After providing your explanation, state your final verdict by strictly following this format: "[[Correct]]" if the response's final answer is correct and "[[Incorrect]]" if it is not.

\vspace{0.5em}
\textbf{[User Question]} \\
\{question\}

\vspace{0.5em}
\textbf{[The Start of Assistant’s Answer]} \\
\{answer\_a\} \\
\textbf{[The End of Assistant’s Answer]}
\end{tcolorbox}
\caption{Prompt templates for CoT answer judgment in the MMLU Pro and GSM-Symbolic datasets}
\label{fig:MMLU_cot_prompt}
\end{figure*}

\begin{figure*}[tp]
\centering
\begin{tcolorbox}[width=\textwidth]
\small
\textbf{[System]} \\
Please act as an impartial meta-judge and evaluate the decision made by an AI judge on the following user question. The user question is a comparative evaluation task, asking to judge which of two AI assistant responses is better based on correctness and helpfulness. \\
\\
The AI judge has already made a selection between Assistant A and Assistant B as the better response, using the following verdict format:  \\
- "[[A]]" if Assistant A is better   \\
- "[[B]]" if Assistant B is better  \\
- "[[C]]" if it is a tie \\

Your task is to independently perform the same comparison step-by-step and determine whether you agree with the AI judge's choice. Focus solely on verifying the correctness of the AI judge's final decision.
\\

After completing your evaluation, provide a concise explanation of your reasoning. Be as objective as possible. Finally, state your verdict in the following format: "**[[Correct]]**" if the AI judge's selection is correct, or "**[[Incorrect]]**" if it is not. \\

\vspace{0.5em}
\textbf{User Question:} \\
\{question\} \\

\vspace{0.5em}
\textbf{[The Start of AI Judge's Selected Verdict]} \\
\{response\} \\
\textbf{[The End of AI Judge's Selected Verdict]}
\end{tcolorbox}
\caption{Prompt templates for CoT answer judgment in the Chatbot Arena and MT-Bench datasets}
\label{fig:chatbot_arena_cot_prompt}
\end{figure*}

% ---------------------------------------- CoT Prompt ----------------------------------------------

% ---------------------------------------- ref Prompt ----------------------------------------------
\begin{figure*}[tp]
\centering
\begin{tcolorbox}[width=\textwidth]
\small
\textbf{[System]} \\
Please act as an impartial judge and evaluate the quality of the response provided by an AI assistant to the user question displayed below. You will be given a reference answer and the assistant's answer. Your evaluation should focus solely on the correctness of the assistant's final answer. Begin by independently solving the user question step-by-step to verify the correctness of the response, and compare your final answer with both the reference answer and the assistant's final answer. Provide a brief explanation of your judgment, highlighting any differences and their significance. Be as objective as possible. After providing your explanation, state your final verdict by strictly following this format: "[[Correct]]" if the response's final answer is correct and "[[Incorrect]]" if it is not.

\vspace{0.5em}
\textbf{[User Question]} \\
\{question\}

\vspace{0.5em}
\textbf{[The Start of Reference Answer]} \\
\{ref\_answer\} \\
\textbf{[The End of Reference Answer]}

\vspace{0.5em}
\textbf{[The Start of Assistant’s Answer]} \\
\{answer\_a\} \\
\textbf{[The End of Assistant’s Answer]}
\end{tcolorbox}
\caption{Prompt templates for self-reference-guided answer judgment in the MMLU Pro and GSM-Symbolic datasets}
\label{fig:MMLU_ref_guided_prompt}
\end{figure*}

\begin{figure*}[tp]
\centering
\begin{tcolorbox}[width=\textwidth]
\small
\textbf{[System]} \\
Please act as an impartial meta-judge and evaluate the decision made by an AI judge on the following user question. The user question is a comparative evaluation task, asking to judge which of two AI assistant responses is better based on correctness and helpfulness. \\

The AI judge has already made a selection between Assistant A and Assistant B as the better response, using the following verdict format: \\
- "[[A]]" if Assistant A is better  \\
- "[[B]]" if Assistant B is better  \\
- "[[C]]" if it is a tie \\

You will also be given a reference answer to the same user question. Your task is to independently perform the same comparison step-by-step and determine whether you agree with the AI judge's choice. Use the reference answer to guide your reasoning and verification, but base your decision on whether the AI judge's final choice was justified given the relative correctness and helpfulness of the two assistant responses. \\

After completing your evaluation, provide a concise explanation of your reasoning. Be as objective as possible. Finally, state your verdict in the following format: "**[[Correct]]**" if the AI judge's selection is correct, or "**[[Incorrect]]**" if it is not.

\vspace{0.5em}
\textbf{User Question:} \\
\{question\} \\

\vspace{0.5em}
\textbf{[The Start of Reference Answer]} \\
\{ref\_answer\} \\
\textbf{[The End of Reference Answer]} \\

\vspace{0.5em}
\textbf{[The Start of AI Judge's Selected Verdict]} \\
\{response\} \\
\textbf{[The End of AI Judge's Selected Verdict]}
\end{tcolorbox}
\caption{Prompt templates for self-reference-guided answer judgment in the Chatbot Arena and MT-Bench subtasks datasets.}
\label{fig:chatbot_arena_ref_guided_prompt}
\end{figure*}

% \begin{figure*}[tp]
% \centering
% \begin{tcolorbox}[width=\textwidth]
% \small

% \end{tcolorbox}
% \caption{Example of Self-Reference-Guided Judgment on a Math Problem}
% \label{fig:math_example_ref_guided_prompt}
% \end{figure*}

% ---------------------------------------- ref Prompt ----------------------------------------------

%% --------------------------- Analysis 1 figure --------------------------------------
\section{Full Results on All Datasets}
\label{appendix:full_results}
Due to space constraints in the main paper, we presented results on only three representative datasets: MMLU Pro, GSM8K, and MT-Bench Writing. Here, we provide the complete experimental results across all seven datasets used in our study, following the same structure as the main paper.

\subsection{Dataset-Level Observations}
\label{appendix:dataset_level}
\begin{figure*}[htbp]
    \centering
    \includegraphics[width=0.95\textwidth]{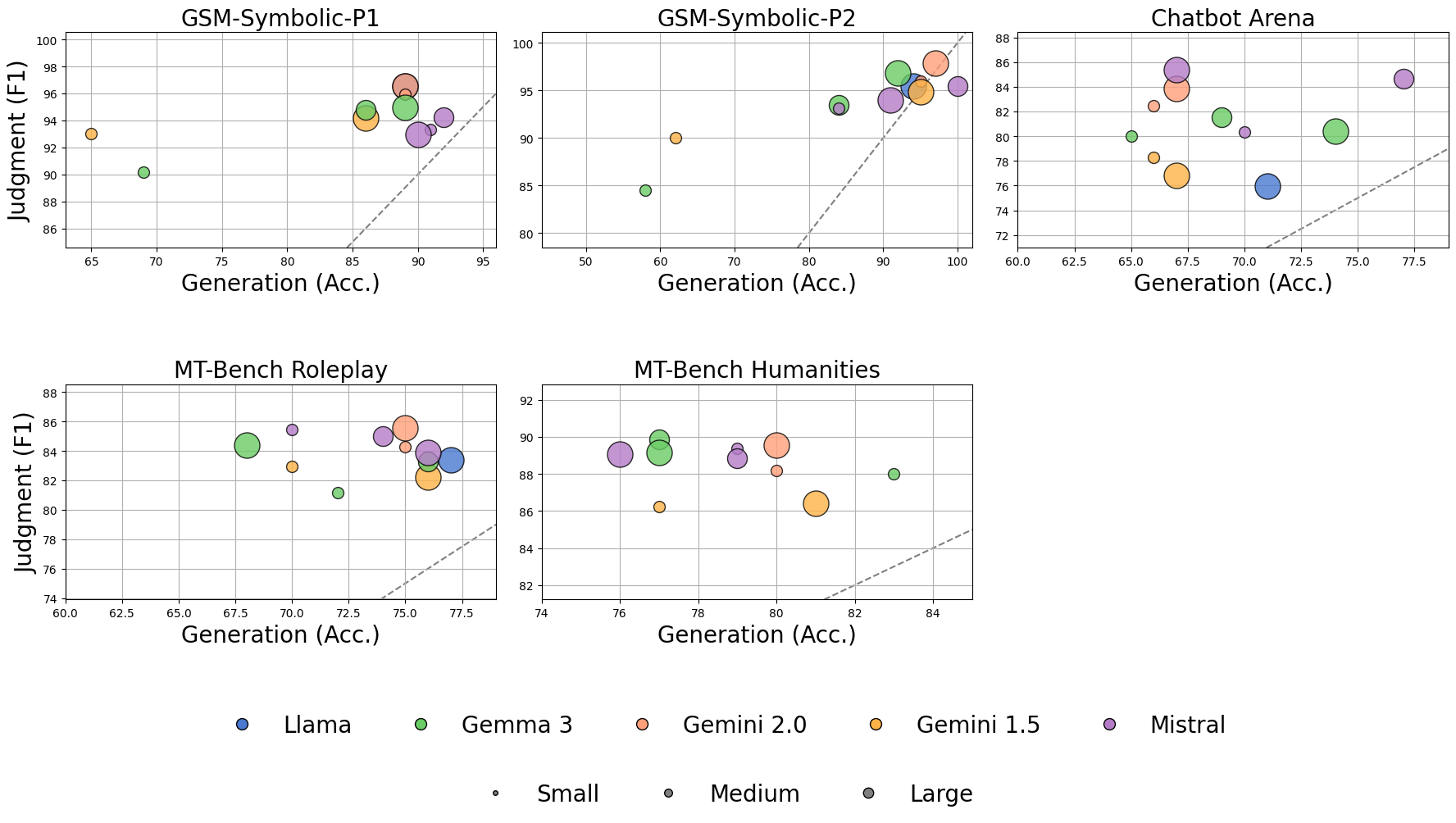} % 圖片文件的路徑
    \caption{Relationship between the capabilities of answer generation (measured by accuracy) and answer judgment (measured by F1 score) across five datasets: GSM-Symbolic-P1, GSM-Symbolic-P2, Chatbot Arena, MT-Bench Roleplay, and MT-Bench Humanities. Each subplot corresponds to one dataset. Different colors represent different model series, and the size of each circle reflects the relative size of models within the same series.}
    \label{fig:other_dataset_capability_comparison}
\end{figure*}

Figure~\ref{fig:other_dataset_capability_comparison} extends our dataset-level analysis to the remaining four datasets. For GSM-Symbolic-P1 and GSM-Symbolic-P2, we observe the same positive correlation trend between answer generation capability and answer judgment capability as seen in the main paper. However, this trend is less pronounced in Chatbot Arena, MT-Bench Humanities, and MT-Bench Roleplay. This variability across datasets underscores our argument that dataset-level correlations alone provide an incomplete picture of the relationship between these two capabilities.
Despite these differences, Tables~\ref{tab:evaluation performance on J+ J- 2} and~\ref{tab:evaluation performance on J+ J- 3} consistently show that performance on $D_{J+}$ outperforms $D_{J-}$ across most models and datasets. This aligns with our hypothesis in the main paper that this phenomenon likely stems from LLMs' tendency to predict answers as correct, rather than from a strong intrinsic correlation between generation and judgment abilities.

\begin{table*}[htbp]
\centering
\small
\setlength{\tabcolsep}{4pt}
\renewcommand{\arraystretch}{1}
\begin{tabular}{lccccccccc}
\hline
\multirow{2}{*}{\textbf{Judge model}}
 & \multicolumn{3}{c}{\textbf{GSM-Symbolic-P1}}
 & \multicolumn{3}{c}{\textbf{GSM-Symbolic-P2}}
 & \multicolumn{3}{c}{\textbf{Chatbot Arena}}
 \\
\cline{2-10}
 & \Checkmark & \XSolidBrush & $\Delta$ & \Checkmark & \XSolidBrush & $\Delta$ & \Checkmark & \XSolidBrush & $\Delta$ \\
\hline
Llama 3.1 405B & \textbf{97.79} & 83.72 & \textcolor{blue}{14.07} & \textbf{96.44} & 70.59 & \textcolor{blue}{25.85} & \textbf{85.56} & 43.40 & \textcolor{blue}{42.16} \\
Gemini 2.0 Flash & \textbf{98.68} & 66.67 & \textcolor{blue}{32.01} & \textbf{97.89} & 100.00 & \textcolor{red}{-2.11} & \textbf{92.53} & 60.47 & \textcolor{blue}{32.06} \\
Gemini 2.0 Flash Lite & \textbf{98.45} & 69.77 & \textcolor{blue}{28.68} & \textbf{96.97} & 66.67 & \textcolor{blue}{30.30} & \textbf{91.71} & 59.31 & \textcolor{blue}{32.40} \\
Gemini 1.5 Flash & \textbf{96.80} & 68.18 & \textcolor{blue}{28.62} & \textbf{94.89} & 94.74 & \textcolor{blue}{0.15} & \textbf{86.16} & 49.54 & \textcolor{blue}{36.62} \\
Gemini 1.5 Flash 8B & \textbf{96.34} & 85.14 & \textcolor{blue}{11.20} & \textbf{91.45} & 87.07 & \textcolor{blue}{4.38} & \textbf{86.90} & 54.84 & \textcolor{blue}{32.06} \\
Gemma 3 4B & \textbf{95.70} & 72.41 & \textcolor{blue}{23.29} & \textbf{91.11} & 72.48 & \textcolor{blue}{18.63} & \textbf{88.55} & 59.42 & \textcolor{blue}{29.13} \\
Gemma 3 12B & \textbf{96.67} & 75.56 & \textcolor{blue}{21.11} & \textbf{95.36} & 80.00 & \textcolor{blue}{15.36} & \textbf{89.20} & 58.82 & \textcolor{blue}{30.38} \\
Gemma 3 27B & \textbf{97.40} & 60.61 & \textcolor{blue}{36.79} & \textbf{97.04} & 94.12 & \textcolor{blue}{2.92} & \textbf{88.04} & 52.94 & \textcolor{blue}{35.10} \\
Mistral Small 3.1 & \textbf{94.78} & 70.97 & \textcolor{blue}{23.81} & \textbf{93.83} & 86.96 & \textcolor{blue}{6.87} & \textbf{88.71} & 53.91 & \textcolor{blue}{34.80} \\
Mistral Medium 3 & \textbf{95.58} & 73.33 & \textcolor{blue}{22.25} & \textbf{95.50} & 0.00 & \textcolor{blue}{95.50} & \textbf{92.66} & 48.28 & \textcolor{blue}{44.38} \\
Mistral Large 2 & \textbf{94.96} & 66.67 & \textcolor{blue}{28.29} & \textbf{95.17} & 81.08 & \textcolor{blue}{14.09} & \textbf{92.96} & 65.15 & \textcolor{blue}{27.81} \\
\hline
\end{tabular}
\caption{Answer judgment performance across different models and datasets. \Checkmark represents $\mathcal{D}_{J+}$, and \XSolidBrush represents $\mathcal{D}_{J-}$. \(\Delta\) denotes the gap between the performance of $\mathcal{D}_{J+}$ and $\mathcal{D}_{J-}$ for the same model and dataset. If the performance of $\mathcal{D}_{J+}$ is higher, it is marked in {\color{blue}{blue}}; otherwise, it is marked in {\color{red}{red}}.}
\label{tab:evaluation performance on J+ J- 2}
\end{table*}

\begin{table*}[htbp]
\centering
\small
\setlength{\tabcolsep}{4pt}
\renewcommand{\arraystretch}{1}
\begin{tabular}{lcccccc}
\hline
\multirow{2}{*}{\textbf{Judge model}}
 & \multicolumn{3}{c}{\textbf{MT-Bench Humanities}}
 & \multicolumn{3}{c}{\textbf{MT-Bench Roleplay}}
 \\
\cline{2-7}
 & \Checkmark & \XSolidBrush & $\Delta$ & \Checkmark & \XSolidBrush & $\Delta$ \\
\hline
Llama 3.1 405B & \textbf{99.78} & 6.90 & \textcolor{blue}{92.88} & \textbf{93.98} & 21.62 & \textcolor{blue}{72.36} \\
Gemini 2.0 Flash & \textbf{97.48} & 10.53 & \textcolor{blue}{86.95} & \textbf{94.76} & 47.42 & \textcolor{blue}{47.34} \\
Gemini 2.0 Flash Lite & \textbf{96.67} & 18.75 & \textcolor{blue}{77.92} & \textbf{92.31} & 50.51 & \textcolor{blue}{41.80} \\
Gemini 1.5 Flash & \textbf{96.11} & 0.00 & \textcolor{blue}{96.11} & \textbf{91.58} & 32.00 & \textcolor{blue}{59.58} \\
Gemini 1.5 Flash 8B & \textbf{95.92} & 0.00 & \textcolor{blue}{95.92} & \textbf{94.06} & 41.90 & \textcolor{blue}{52.16} \\
Gemma 3 4B & \textbf{98.45} & 8.16 & \textcolor{blue}{90.29} & \textbf{90.91} & 46.30 & \textcolor{blue}{44.61} \\
Gemma 3 12B & \textbf{99.77} & 26.67 & \textcolor{blue}{73.10} & \textbf{92.46} & 37.50 & \textcolor{blue}{54.96} \\
Gemma 3 27B & \textbf{99.54} & 19.18 & \textcolor{blue}{80.36} & \textbf{95.85} & 47.93 & \textcolor{blue}{47.92} \\
Mistral Small 3.1 & \textbf{99.08} & 3.33 & \textcolor{blue}{95.75} & \textbf{98.51} & 32.32 & \textcolor{blue}{66.19} \\
Mistral Medium 3 & \textbf{99.55} & 3.28 & \textcolor{blue}{96.27} & \textbf{95.10} & 40.86 & \textcolor{blue}{54.24} \\
Mistral Large 2 & \textbf{99.77} & 8.96 & \textcolor{blue}{90.81} & \textbf{95.37} & 20.51 & \textcolor{blue}{74.86} \\
\hline
\end{tabular}
\caption{Answer judgment performance across different models and datasets. \Checkmark represents $\mathcal{D}_{J+}$, and \XSolidBrush represents $\mathcal{D}_{J-}$. \(\Delta\) denotes the gap between the performance of $\mathcal{D}_{J+}$ and $\mathcal{D}_{J-}$ for the same model and dataset. If the performance of $\mathcal{D}_{J+}$ is higher, it is marked in {\color{blue}{blue}}; otherwise, it is marked in {\color{red}{red}}.}
\label{tab:evaluation performance on J+ J- 3}
\end{table*}

\begin{table*}[htbp]
    \centering
    \small
    \begin{tabular}{lcccccc}
        \hline
        \textbf{Judge Model} & \textbf{\makecell{GSM-\\Symbolic-P1}} & \textbf{\makecell{GSM-\\Symbolic-P2}} & \textbf{\makecell{Chatbot \\Arena}} & \textbf{\makecell{MT-Bench\\Humanities}} & \textbf{\makecell{MT-Bench\\Roleplay}} & \textbf{\makecell{Avg.}} \\
        \hline
        Mistral Large 2 & 10.00\% & 7.67\% & 22.33\% & 17.33\% & 24.66\% & 16.40\% \\
        Gemini 2.0 Flash Lite & 4.00\% & 5.00\% & 28.33\% & 18.00\% & 26.33\% & 16.33\% \\
        Mistral Medium 3 & 7.67\% & 5.34\% & 20.67\% & 17.00\% & 21.66\% & 14.47\% \\
        Mistral Small 3.1 & 9.34\% & 7.67\% & 19.33\% & 13.66\% & 21.66\% & 14.33\% \\
        Gemma 3 27B & 4.34\% & 4.00\% & 16.33\% & 19.33\% & 23.66\% & 13.53\% \\
        Gemma 3 12B & 4.67\% & 5.00\% & 16.00\% & 18.00\% & 14.00\% & 11.53\% \\
        Gemini 2.0 Flash & 2.34\% & 1.67\% & 19.00\% & 13.33\% & 20.66\% & 11.40\% \\
        Llama 3.1 405B & 4.00\% & 2.34\% & 11.00\% & 16.66\% & 22.66\% & 11.33\% \\
        Gemini 1.5 Flash 8B & 7.00\% & 7.67\% & 13.33\% & 9.66\% & 18.66\% & 11.26\% \\
        Gemma 3 4B & 2.00\% & -3.66\% & 15.00\% & 14.66\% & 17.33\% & 9.07\% \\
        Gemini 1.5 Flash & 0.00\% & 0.67\% & 2.33\% & 6.33\% & 14.33\% & 4.73\% \\
        \hline
    \end{tabular}%
    \caption{Model overconfidence across datasets, showing the difference between percentage of samples predicted as correct and percentage actually correct. Higher values indicate greater bias toward predicting correctness. The average (rightmost column) is computed as a weighted mean across all datasets based on sample count.}
    \label{tab:overconfidence 2}
\end{table*}

\begin{figure*}[htbp]
    \centering
    \includegraphics[width=1\textwidth, height=7.5cm]{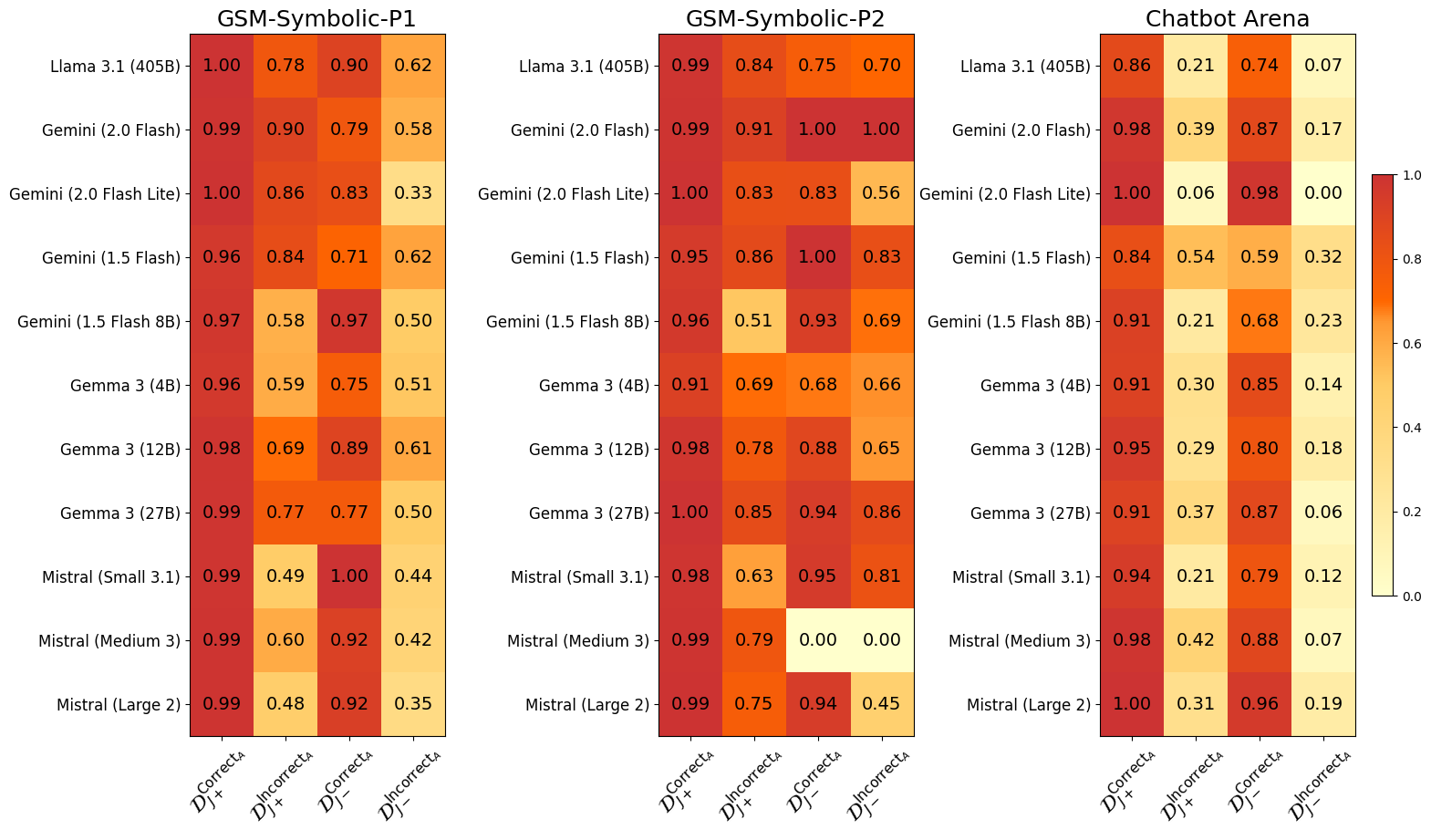} % 圖片文件的路徑
    \caption{Heatmap Visualization of Evaluation Performance across Datasets. This figure illustrates the integration of the judge model's answer generation capabilities with the labels of evaluation questions across four datasets. Each dataset is represented in one of four subfigures, with subsets $\mathcal{D}_{J+}^{\text{Correct}_A}$, $\mathcal{D}_{J+}^{\text{Incorrect}_A}$, $\mathcal{D}_{J-}^{\text{Correct}_A}$, and $\mathcal{D}_{J-}^{\text{Incorrect}_A}$ displayed from left to right, showing the evaluation accuracy variations under different conditions.}
    \label{fig:full_heatmap_2}
\end{figure*}

\begin{figure*}[htbp]
    \centering
    \includegraphics[width=0.7\textwidth, height=7.5cm]{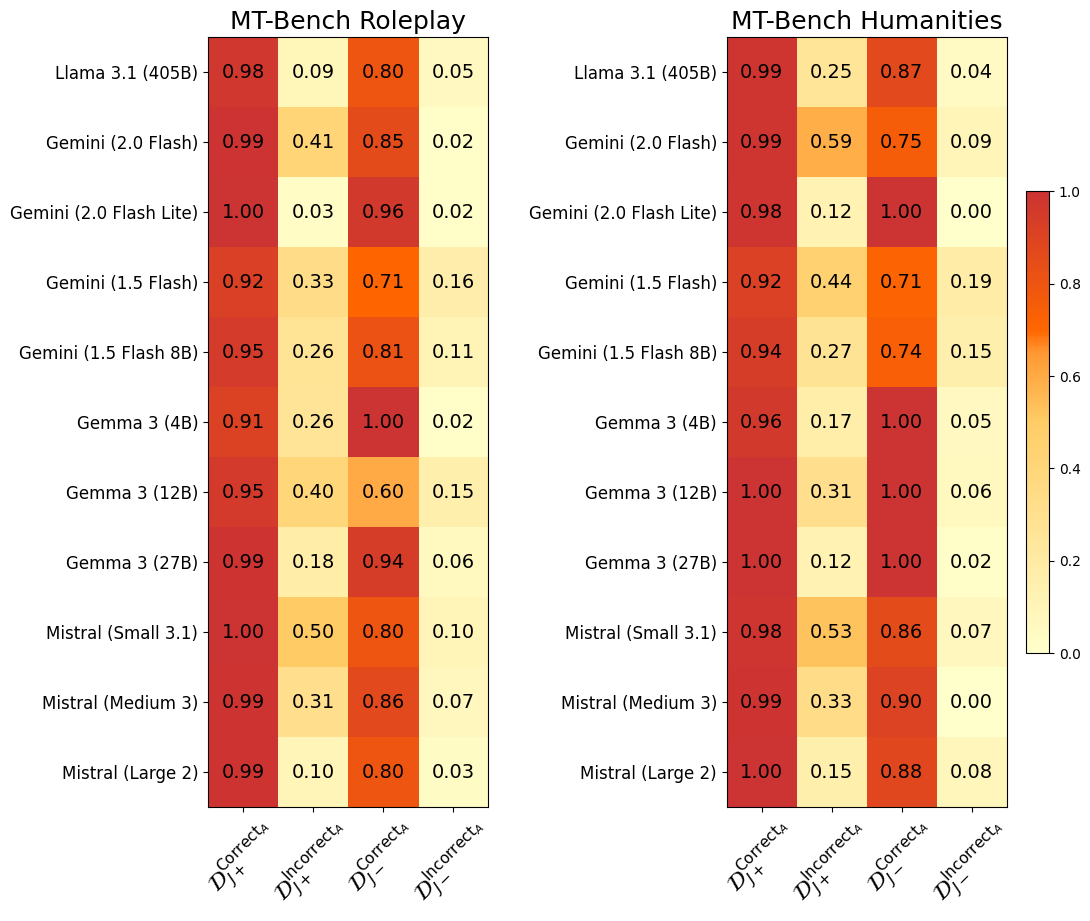} % 圖片文件的路徑
    \caption{Heatmap Visualization of Evaluation Performance across Datasets. This figure illustrates the integration of the judge model's answer generation capabilities with the labels of evaluation questions across four datasets. Each dataset is represented in one of four subfigures, with subsets $\mathcal{D}_{J+}^{\text{Correct}_A}$, $\mathcal{D}_{J+}^{\text{Incorrect}_A}$, $\mathcal{D}_{J-}^{\text{Correct}_A}$, and $\mathcal{D}_{J-}^{\text{Incorrect}_A}$ displayed from left to right, showing the evaluation accuracy variations under different conditions.}
    \label{fig:full_heatmap_3}
\end{figure*}

\subsection{In-Depth Dataset-Level Analysis}
\label{appendix:finer_grained}

Figures~\ref{fig:full_heatmap_2} and~\ref{fig:full_heatmap_3} present the finer-grained analysis for the remaining datasets, where we partition each dataset into four subsets based on the judge model's answer correctness ($J+$ or $J-$) and the agent model's answer correctness (Correct$_A$ or Incorrect$_A$).
For Chatbot Arena, MT-Bench Humanities, and MT-Bench Roleplay, we observe patterns similar to those shown for MMLU Pro and MT-Bench Writing in the main paper. Specifically, LLMs consistently perform best on $\mathcal{D}_{J+}^{\text{Correct}_A}$ and $\mathcal{D}_{J-}^{\text{Correct}_A}$ subsets, further confirming our finding that LLM-as-Judge performance is strongly influenced by the correctness of the agent's response, rather than by the judge model's ability to answer the question correctly.

For GSM-Symbolic-P1 and GSM-Symbolic-P2, the results mirror those of GSM8K. This similarity likely stems from the high answer generation accuracy across models for these mathematical reasoning tasks, resulting in very few instances falling into the Incorrect$_J$ and Incorrect$_A$ categories, which limits the conclusiveness of analyses for these particular subsets.
Table~\ref{tab:overconfidence 2} extends our analysis of prediction behavior across the remaining datasets. Consistent with our findings in the main paper, we observe that most LLMs exhibit a strong bias toward predicting answers as \textit{Correct}. This systematic overconfidence manifests across all datasets, though with varying degrees of intensity. The consistency of this pattern supports our hypothesis that the performance gap between $D_{J+}$ and $D_{J-}$ is heavily influenced by this prediction bias rather than by an intrinsic correlation between generation and judgment abilities.

\subsection{Instance-Level Analysis}
\label{appendix:instance_level}

Tables~\ref{tab:partial_correlation 2} and~\ref{tab:partial_correlation 3} present the partial correlation results for the remaining five datasets. Consistent with our main findings, these results show that the partial correlations between answer generation and judgment abilities remain low across most cases, with the majority of values falling below 0.3. This further strengthens our conclusion that these two capabilities are only weakly correlated when controlling for the correctness of the agent's response.

A notable observation is that \texttt{Mistral Medium 3} exhibits a partial correlation of 0 on GSM-Symbolic-P2. This is directly attributable to the model achieving 100\% accuracy in answer generation on this dataset, resulting in no instances where the judge model answers incorrectly ($J-$). This edge case illustrates the limitations of correlation analysis when performance approaches perfection on either task, but does not contradict our broader findings about the weak correlation between these capabilities under normal circumstances.

\begin{table}[htbp]
    \centering
    \resizebox{\columnwidth}{!}{%
    \begin{tabular}{lccc}
        \toprule
     \textbf{Judge Model} & \textbf{\makecell{GSM-\\Symbolic-P1}} & \textbf{\makecell{GSM-\\Symbolic-P2}} & \textbf{\makecell{Chatbot\\Arena}} \\
        \midrule
        Llama 3.1 405B & \cellcolor{red!20}{0.1822} & \cellcolor{red!20}{0.1816} & \cellcolor{red!20}{0.1415} \\
        Gemini 2.0 Flash & \cellcolor{blue!20}{0.3526} & \cellcolor{red!20}{-0.0609} & \cellcolor{red!20}{0.2222} \\
        Gemini 2.0 Flash Lite & \cellcolor{blue!20}{0.4143} & \cellcolor{red!20}{0.2239} & \cellcolor{red!20}{0.1626} \\
        Gemini 1.5 Flash & \cellcolor{red!20}{0.2791} & \cellcolor{red!20}{-0.0153} & \cellcolor{red!20}{0.2376} \\
        Gemini 1.5 Flash 8B & \cellcolor{red!20}{0.0313} & \cellcolor{red!20}{-0.0508} & \cellcolor{red!20}{0.1690} \\
        Gemma 3 4B & \cellcolor{red!20}{0.2325} & \cellcolor{red!20}{0.2029} & \cellcolor{red!20}{0.1209} \\
        Gemma 3 12B & \cellcolor{red!20}{0.1113} & \cellcolor{red!20}{0.1432} & \cellcolor{red!20}{0.1777} \\
        Gemma 3 27B & \cellcolor{red!20}{0.2852} & \cellcolor{red!20}{0.0529} & \cellcolor{red!20}{0.1922} \\
        Mistral Small 3.1 & \cellcolor{red!20}{0.0231} & \cellcolor{red!20}{-0.0989} & \cellcolor{red!20}{0.1689} \\
        Mistral Medium 3 & \cellcolor{red!20}{0.1338} & \cellcolor{red!20}{0.0000} & \cellcolor{blue!20}{0.3291} \\
        Mistral Large 2 & \cellcolor{red!20}{0.1127} & \cellcolor{red!20}{0.1666} & \cellcolor{red!20}{0.1300} \\
        \bottomrule
    \end{tabular}%
    }
   \caption{Partial correlation between answer generation and judgment capabilities across models and datasets, controlling for the correctness of evaluated responses. Weak and moderate correlations are highlighted with \colorbox{red!30}{red} and \colorbox{blue!30}{purple} backgrounds, respectively.}
    \label{tab:partial_correlation 2}
\end{table}

\begin{table}[htbp]
    \centering
    \resizebox{\columnwidth}{!}{%
    \begin{tabular}{lcc}
        \toprule
        \textbf{Judge Model} & \textbf{\makecell{MT-Bench\\Humanities}} & \textbf{\makecell{MT-Bench\\Roleplay}} \\
        \midrule
        Llama 3.1 405B & \cellcolor{red!20}{0.2519} & \cellcolor{red!20}{0.1316} \\
        Gemini 2.0 Flash & \cellcolor{blue!20}{0.4404} & \cellcolor{blue!20}{0.4085} \\
        Gemini 2.0 Flash Lite & \cellcolor{red!20}{0.0994} & \cellcolor{red!20}{0.0960} \\
        Gemini 1.5 Flash & \cellcolor{red!20}{0.2043} & \cellcolor{red!20}{0.1927} \\
        Gemini 1.5 Flash 8B & \cellcolor{red!20}{0.1813} & \cellcolor{red!20}{0.1836} \\
        Gemma 3 4B & \cellcolor{red!20}{0.0613} & \cellcolor{red!20}{0.0662} \\
        Gemma 3 12B & \cellcolor{red!20}{0.1859} & \cellcolor{blue!20}{0.3448} \\
        Gemma 3 27B & \cellcolor{red!20}{0.1243} & \cellcolor{red!20}{0.1465} \\
        Mistral Small 3.1 & \cellcolor{blue!20}{0.3558} & \cellcolor{blue!20}{0.3762} \\
        Mistral Medium 3 & \cellcolor{blue!20}{0.3463} & \cellcolor{red!20}{0.2808} \\
        Mistral Large 2 & \cellcolor{red!20}{0.2095} & \cellcolor{red!20}{0.2057} \\
        \bottomrule
    \end{tabular}%
    }
   \caption{Partial correlation between answer generation and judgment capabilities across models and datasets, controlling for the correctness of evaluated responses. Weak and moderate correlations are highlighted with \colorbox{red!30}{red} and \colorbox{blue!30}{purple} backgrounds, respectively.}
    \label{tab:partial_correlation 3}
\end{table}

\subsection{Self-Reference-Guided Results}
\label{appendix:self_reference}

Tables~\ref{tab:partial_correlation_diff 2} and~\ref{tab:partial_correlation_diff 3} demonstrate the effectiveness of our self-reference-guided method across the remaining five datasets. The results strongly reinforce our findings from the main paper: after applying this method, the partial correlations between answer generation and judgment capabilities increase dramatically in most cases, with values typically exceeding 0.6. Most improvements show gains of over 0.4 in correlation strength compared to the standard CoT approach.

This consistent performance across diverse datasets, including mathematical reasoning (GSM-Symbolic-P1, GSM-Symbolic-P2), open-ended dialogue evaluation (Chatbot Arena, MT-Bench Roleplay), and humanities-focused dialogue (MT-Bench Humanities), validates the generalizability of our self-reference-guided approach. The method is effective across different model architectures and task types, confirming that using a model's own answers as references reliably aligns its generation and judgment capabilities.

% This consistent performance across diverse datasets—spanning mathematical reasoning (GSM-Symbolic-P1, GSM-Symbolic-P2), open-ended dialogue evaluation (Chatbot Arena, MT-Bench Roleplay), and humanities-focused dialogue (MT-Bench Humanities)—validates the generalizability of our self-reference-guided approach. The method proves effective across various model architectures and task types, confirming that leveraging a model's own answers as references reliably aligns its generation and judgment capabilities.

\subsection{Error Analysis}
\label{sec:appendix_analysis}

\begin{figure*}[htbp]
    \centering
    \includegraphics[width=1\textwidth, height=7.5cm]{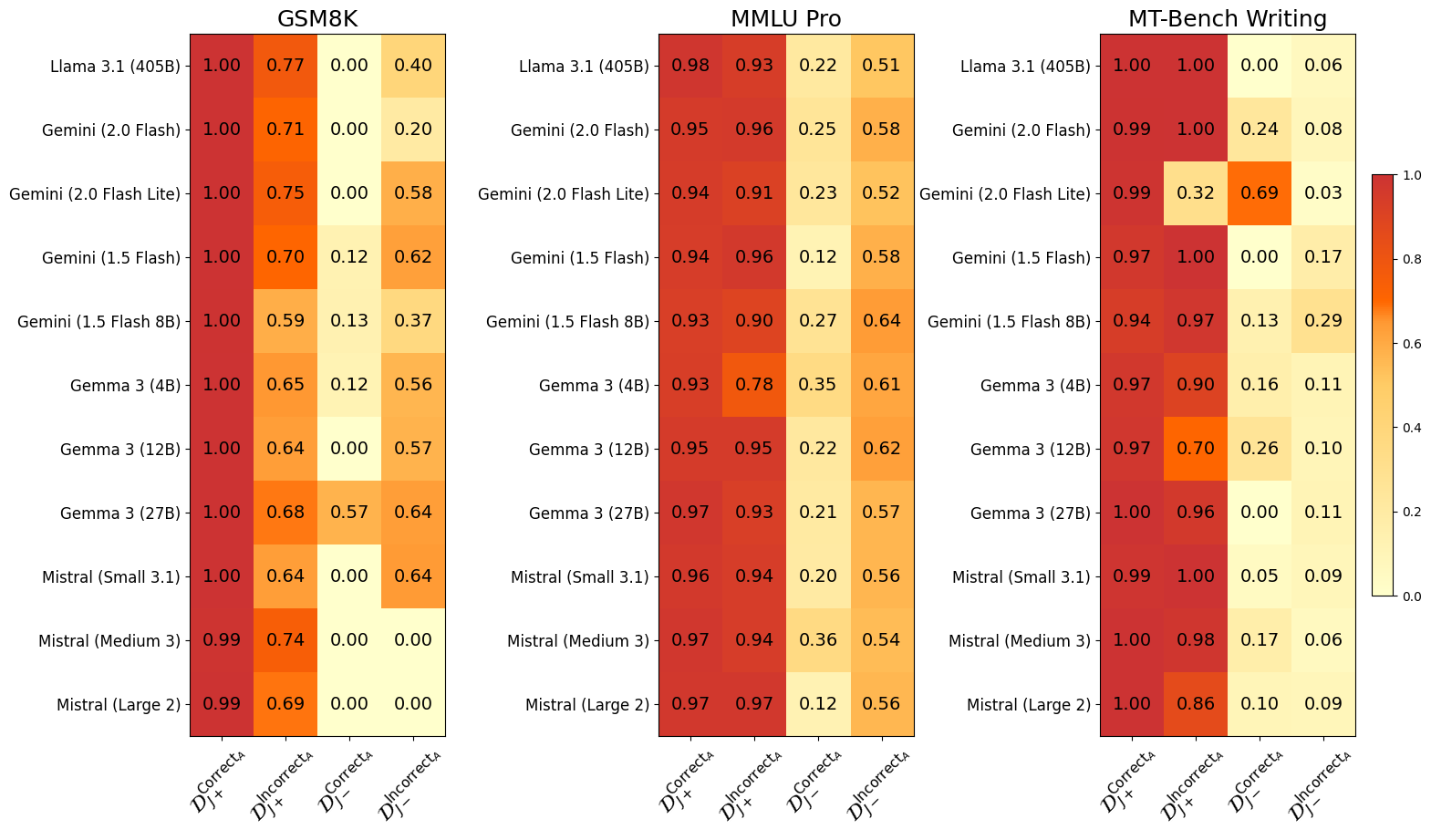} % 圖片文件的路徑
    \caption{Heatmap Visualization of Evaluation Performance across Datasets under the Self-Reference Guided Method. This figure illustrates the integration of the judge model's answer generation capabilities with the labels of evaluation questions across four datasets. Each dataset is represented in one of four subfigures, with subsets $\mathcal{D}_{J+}^{\text{Correct}_A}$, $\mathcal{D}_{J+}^{\text{Incorrect}_A}$, $\mathcal{D}_{J-}^{\text{Correct}_A}$, and $\mathcal{D}_{J-}^{\text{Incorrect}_A}$ displayed from left to right, showing the evaluation accuracy variations under different conditions.}
    \label{fig:full_heatmap_ref_1}
\end{figure*}

\begin{figure*}[htbp]
    \centering
    \includegraphics[width=1\textwidth, height=7.5cm]{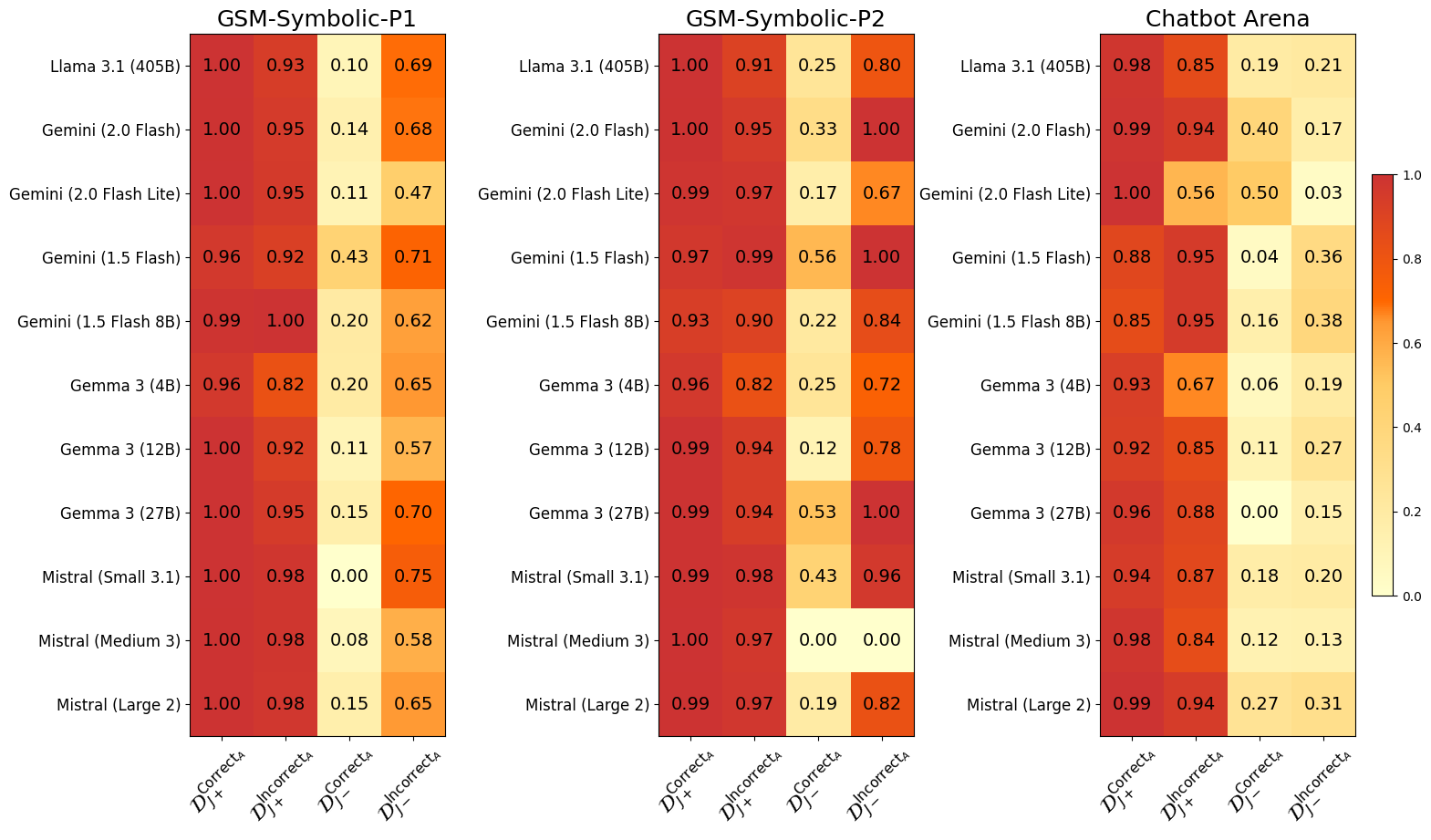} % 圖片文件的路徑
    \caption{Heatmap Visualization of Evaluation Performance across Datasets under the Self-Reference Guided Method. This figure illustrates the integration of the judge model's answer generation capabilities with the labels of evaluation questions across four datasets. Each dataset is represented in one of four subfigures, with subsets $\mathcal{D}_{J+}^{\text{Correct}_A}$, $\mathcal{D}_{J+}^{\text{Incorrect}_A}$, $\mathcal{D}_{J-}^{\text{Correct}_A}$, and $\mathcal{D}_{J-}^{\text{Incorrect}_A}$ displayed from left to right, showing the evaluation accuracy variations under different conditions.}
    \label{fig:full_heatmap_ref_2}
\end{figure*}

\begin{figure*}[htbp]
    \centering
    \includegraphics[width=0.7\textwidth, height=7.5cm]{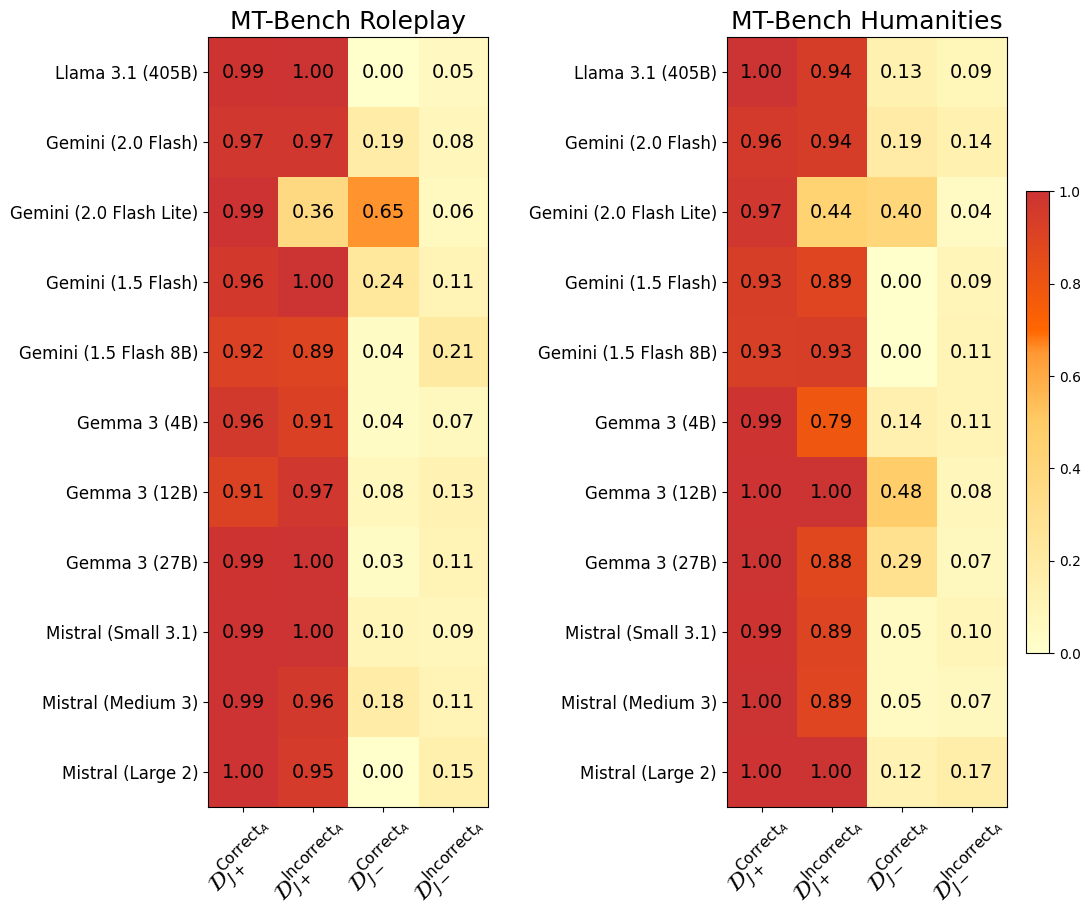} % 圖片文件的路徑
    \caption{Heatmap Visualization of Evaluation Performance across Datasets under the Self-Reference Guided Method. This figure illustrates the integration of the judge model's answer generation capabilities with the labels of evaluation questions across four datasets. Each dataset is represented in one of four subfigures, with subsets $\mathcal{D}_{J+}^{\text{Correct}_A}$, $\mathcal{D}_{J+}^{\text{Incorrect}_A}$, $\mathcal{D}_{J-}^{\text{Correct}_A}$, and $\mathcal{D}_{J-}^{\text{Incorrect}_A}$ displayed from left to right, showing the evaluation accuracy variations under different conditions.}
    \label{fig:full_heatmap_ref_3}
\end{figure*}

A significant challenge for LLM-as-Judge frameworks is that models are often biased towards positive confirmation, performing better when identifying correct answers than incorrect ones. In this section, we analyze how our self-reference-guided method impacts this behavior. Following the finer-grained observation methodology from Section \ref{sec:4_2}, we generated performance heatmaps for the self-reference-guided method, which can be seen in Figure ~\ref{fig:full_heatmap_ref_1},  ~\ref{fig:full_heatmap_ref_2} and  ~\ref{fig:full_heatmap_ref_3}.

\paragraph{Enhancing Error Detection with Correct Knowledge}
The most notable result is the substantial improvement in identifying incorrect answers when the judge model has the correct knowledge ($D_{J+}^{\text{Incorrect}_A}$). As shown in the data for MMLU Pro, the performance in this subset using the self-reference-guided method is exceptionally high, with F1 scores ranging from \textbf{0.78 to 0.97} (mostly above 0.90). This is a stark contrast to the CoT method, where performance on the same subset ranged from 0.46 to 0.75. This demonstrates that the self-reference-guided method helps models more effectively use what they know to identify what is wrong.

\paragraph{A Shift in Judgment Dependency}
This finding reveals a crucial shift in how models perform judgment.
\begin{itemize}
    \item With \textbf{CoT}, performance is primarily dependent on the agent's answer label (i.e., models perform best on the $\mathcal{D}^{\text{Correct}_A}$ subsets).
    \item With \textbf{self-reference}, performance becomes primarily dependent on the judge's own knowledge (i.e., models perform best on the $\mathcal{D}_{J+}$ subsets).
\end{itemize}
This shift suggests that self-reference fundamentally changes the evaluation task from simple label prediction to a process of verification against an internal knowledge base.

\paragraph{Trade-offs and Considerations}
This method is not without trade-offs. While performance on the $\mathcal{D}_{J+}$ subsets improves, performance on the $\mathcal{D}_{J-}$ subsets (where the judge's initial answer is wrong) degrades. This is an expected outcome of the method: a judge model that is confident in its own incorrect answer will use it as a faulty reference, penalizing agent answers that may in fact be correct. This underscores the importance of our primary recommendation: the self-reference-guided method is most reliable when applied to judge models with high generation accuracy in the target domain.

\begin{table*}[htbp]
    \centering
    \small
    % \resizebox{\columnwidth}{!}{%
    \begin{tabular}{lccc}
        \toprule
        \textbf{Judge Model} & \textbf{GSM-Symbolic-P1} & \textbf{GSM-Symbolic-P2} & \textbf{Chatbot Arena} \\
        \midrule
        Llama 3.1 405B & \cellcolor{green!20}{0.7438} (\textcolor{blue}{+0.5616}↑) & \cellcolor{blue!20}{0.4440} (\textcolor{blue}{+0.2624}↑) & \cellcolor{green!20}{0.7214} (\textcolor{blue}{+0.5799}↑) \\
        Gemini 2.0 Flash & \cellcolor{green!20}{0.6554} (\textcolor{blue}{+0.3028}↑) & \cellcolor{red!20}{0.2208} (\textcolor{blue}{+0.2817}↑) & \cellcolor{green!20}{0.7230} (\textcolor{blue}{+0.5008}↑) \\
        Gemini 2.0 Flash Lite & \cellcolor{green!20}{0.7965} (\textcolor{blue}{+0.3822}↑) & \cellcolor{green!20}{0.5524} (\textcolor{blue}{+0.3285}↑) & \cellcolor{green!20}{0.6399} (\textcolor{blue}{+0.4773}↑) \\
        Gemini 1.5 Flash & \cellcolor{blue!20}{0.4480} (\textcolor{blue}{+0.1689}↑) & \cellcolor{red!20}{0.2722} (\textcolor{blue}{+0.2875}↑) & \cellcolor{green!20}{0.6927} (\textcolor{blue}{+0.4551}↑) \\
        Gemini 1.5 Flash 8B & \cellcolor{green!20}{0.7467} (\textcolor{blue}{+0.7154}↑) & \cellcolor{green!20}{0.5679} (\textcolor{blue}{+0.6187}↑) & \cellcolor{green!20}{0.6119} (\textcolor{blue}{+0.4429}↑) \\
        Gemma 3 4B & \cellcolor{green!20}{0.6434} (\textcolor{blue}{+0.4109}↑) & \cellcolor{green!20}{0.5721} (\textcolor{blue}{+0.3692}↑) & \cellcolor{green!20}{0.6963} (\textcolor{blue}{+0.5754}↑) \\
        Gemma 3 12B & \cellcolor{green!20}{0.7200} (\textcolor{blue}{+0.6087}↑) & \cellcolor{green!20}{0.6567} (\textcolor{blue}{+0.5135}↑) & \cellcolor{green!20}{0.6957} (\textcolor{blue}{+0.5180}↑) \\
        Gemma 3 27B & \cellcolor{green!20}{0.6497} (\textcolor{blue}{+0.3645}↑) & \cellcolor{blue!20}{0.4009} (\textcolor{blue}{+0.3480}↑) & \cellcolor{green!20}{0.8182} (\textcolor{blue}{+0.6260}↑) \\
        Mistral Small 3.1 & \cellcolor{green!20}{0.7191} (\textcolor{blue}{+0.6960}↑) & \cellcolor{blue!20}{0.4817} (\textcolor{blue}{+0.5806}↑) & \cellcolor{green!20}{0.7120} (\textcolor{blue}{+0.5431}↑) \\
        Mistral Medium 3 & \cellcolor{green!20}{0.7559} (\textcolor{blue}{+0.6221}↑) & \cellcolor{red!20}{0.0000} (\textcolor{blue}{+0.0000}↑) & \cellcolor{green!20}{0.7736} (\textcolor{blue}{+0.4445}↑) \\
        Mistral Large 2 & \cellcolor{green!20}{0.6923} (\textcolor{blue}{+0.5796}↑) & \cellcolor{green!20}{0.6620} (\textcolor{blue}{+0.4954}↑) & \cellcolor{green!20}{0.7324} (\textcolor{blue}{+0.6024}↑) \\
        \bottomrule
    \end{tabular}%
    % }
    \caption{Partial correlation after applying the {self-reference-guided} judgment method. Improvements over a CoT baseline are shown in \textcolor{blue}{blue} with ↑ arrows. Weak, moderate, and strong correlations are highlighted with \colorbox{red!30}{red}, \colorbox{blue!30}{purple}, and \colorbox{green!30}{green} backgrounds, respectively.}
    \label{tab:partial_correlation_diff 2}
\end{table*}

\begin{table*}[htbp]
    \centering
    \small
    % \resizebox{\columnwidth}{!}{%
    \begin{tabular}{lcc}
        \toprule
        \textbf{Judge Model} & \textbf{MT-Bench Humanities} & \textbf{MT-Bench Roleplay} \\
        \midrule
        Llama 3.1 405B & \cellcolor{green!20}{0.8666} (\textcolor{blue}{+0.6147}↑) & \cellcolor{green!20}{0.9131} (\textcolor{blue}{+0.7815}↑) \\
        Gemini 2.0 Flash & \cellcolor{green!20}{0.6929} (\textcolor{blue}{+0.2525}↑) & \cellcolor{green!20}{0.8299} (\textcolor{blue}{+0.4214}↑) \\
        Gemini 2.0 Flash Lite & \cellcolor{green!20}{0.5356} (\textcolor{blue}{+0.4362}↑) & \cellcolor{blue!20}{0.4518} (\textcolor{blue}{+0.3558}↑) \\
        Gemini 1.5 Flash & \cellcolor{green!20}{0.7059} (\textcolor{blue}{+0.5016}↑) & \cellcolor{green!20}{0.7605} (\textcolor{blue}{+0.5678}↑) \\
        Gemini 1.5 Flash 8B & \cellcolor{green!20}{0.7459} (\textcolor{blue}{+0.5646}↑) & \cellcolor{green!20}{0.6855} (\textcolor{blue}{+0.5019}↑) \\
        Gemma 3 4B & \cellcolor{green!20}{0.7650} (\textcolor{blue}{+0.7037}↑) & \cellcolor{green!20}{0.8349} (\textcolor{blue}{+0.7687}↑) \\
        Gemma 3 12B & \cellcolor{green!20}{0.7328} (\textcolor{blue}{+0.5469}↑) & \cellcolor{green!20}{0.7395} (\textcolor{blue}{+0.3947}↑) \\
        Gemma 3 27B & \cellcolor{green!20}{0.8114} (\textcolor{blue}{+0.6871}↑) & \cellcolor{green!20}{0.8897} (\textcolor{blue}{+0.7432}↑) \\
        Mistral Small 3.1 & \cellcolor{green!20}{0.8649} (\textcolor{blue}{+0.5091}↑) & \cellcolor{green!20}{0.8580} (\textcolor{blue}{+0.4818}↑) \\
        Mistral Medium 3 & \cellcolor{green!20}{0.9029} (\textcolor{blue}{+0.5566}↑) & \cellcolor{green!20}{0.8429} (\textcolor{blue}{+0.5621}↑) \\
        Mistral Large 2 & \cellcolor{green!20}{0.8336} (\textcolor{blue}{+0.6241}↑) & \cellcolor{green!20}{0.8118} (\textcolor{blue}{+0.6061}↑) \\
        \bottomrule
    \end{tabular}%
    % }
    \caption{Partial correlation after applying the \textit{self-reference-guided} judgment method. Improvements over a CoT baseline are shown in \textcolor{blue}{blue} with ↑ arrows. Weak, moderate, and strong correlations are highlighted with \colorbox{red!30}{red}, \colorbox{blue!30}{purple}, and \colorbox{green!30}{green} backgrounds, respectively.}
    \label{tab:partial_correlation_diff 3}
\end{table*}

\end{document}